\documentclass{article}
\usepackage{acra}
\usepackage{graphicx}
\usepackage{subcaption}
\usepackage{placeins}
\usepackage{amsmath}
\usepackage{mathtools}
\usepackage{amsfonts}
\usepackage{booktabs}
\usepackage{array}
\usepackage{balance}

\pdfoutput=1

\newcolumntype{M}[1]{>{\centering\arraybackslash}m{#1}}

\title{Simultaneous Optical Flow and Segmentation (SOFAS) using Dynamic Vision Sensor}
\author{Timo Stoffregen$^{\ast\dagger}$, Lindsay Kleeman$^\ast$ \\ $^\ast$Dept. of Electrical and Computer Systems Engineering \\ Monash University, Australia \\$^\dagger$Australian Centre of Excellence for Robotic Vision\\ timo.stoffregen@monash.edu}
\begin{document}
\bibliographystyle{named}
\maketitle
\begin{abstract}
We present an algorithm (SOFAS) to estimate the optical flow of events generated by a dynamic vision sensor (DVS). Where traditional cameras produce frames at a fixed rate, DVSs produce asynchronous events in response to intensity changes with a high temporal resolution. Our algorithm uses the fact that events are generated by edges in the scene to not only estimate the optical flow but also to simultaneously segment the image into objects which are travelling at the same velocity. This way it is able to avoid the aperture problem which affects other implementations such as Lucas-Kanade. Finally, we show that SOFAS produces more accurate results than traditional optic flow algorithms.
\end{abstract}
\section{Introduction}
%Introduce the DVS, talk about the requiement for optical flow
Dynamic Vision Sensors (DVS), also known as Event Cameras or Neuromorphic Cameras avail robots of an entirely new class of visual information. Where traditional cameras produce image frames at a fixed rate, the pixels of a DVS produce events asynchronously in response to illumination intensity changes, analogously to the way a biological retina generates neural spikes in response to intensity change \cite{Lich2008128x}. The data generated by such a sensor is inherently sparse, since elements of the scene which do not change much in the image such as backgrounds produce little or no data. At the same time elements in the scene which do change are able to receive corresponding ``framerates" in the hundreds of kilohertz \cite{Ni2013,Delbruck2013,Delbruck2015} . This vastly reduces the need for computational power in many classic computer vision applications \cite{brandli2014240}.\par
In computer vision, optic flow has long been a key task in object tracking, image registration, visual odometry, SLAM and other robot navigation techniques \cite{Aires2008Opti}. Computing optic flow is not a trivial task and has been the theme of many research papers in computer vision. While many techniques for optic flow detection exist, the most common implementations are still the original Horn-Schunck \cite{Horn1980Deter} and Lucas-Kanade \cite{Lucas1981Iter} algorithms. However, since most optic flow estimations are based on local computations on a small patch of the image they tend to suffer from the aperture problem \cite{Binder2009}. This states that the motion of one dimensional structures in the image cannot be unambiguously determined if the ends of the stimulus are not visible. Our algorithm avoids the aperture problem by examining the entirety of the structure rather than performing a local estimate.\par
While the suppression of redundancy and high temporal resolution of these data-driven, frameless sensors has presented the possibility of performing demanding visual processing in real time, it also requires a rethink in how the data is processed, since the majority of computer vision algorithms cannot be directly used on event streams. Simply translating existing algorithms to the event-based paradigm often misses the opportunities presented by DVSs. Further, many important recent algorithms accumulate past events to detect image gradients \cite{Ni2012Asynch,Benosman2012Asyn,Benosman2014Even,Kim2014Simu}. While these algorithms allow events to asynchronously alter the current state, by buffering they lose much of the advantage of the high resolution timestamps.\par
Our contribution is to introduce a novel way of computing the optic flow of events produced by a DVS, which exploits the properties of event based data. Our approach is demonstrated to produce more accurate results than traditional optic flow techniques on such data and is able to adapt to evolving velocities and form changes of the underlying structures. Further, our approach is able to segment events by the structures generating them, in the process avoiding the aperture problem entirely. 
%\FloatBarrier
\section{Related Work}
Optical flow using event based sensors is a topic that has been visited frequently, resulting in many different algorithms. A thorough review and evaluation of these was performed in \cite{Rueckauer2014Eval}. Notably all of the dense flow algorithms require the computation of the first derivatives of some image quantity, be it gradients of the luminance intensities or gradients of the timestamps of buffered events. Our method does not require the use of any derivatives.\par
\begin{figure}
    \centering
    \begin{subfigure}[b]{0.48\columnwidth}
        \includegraphics[width=\textwidth]{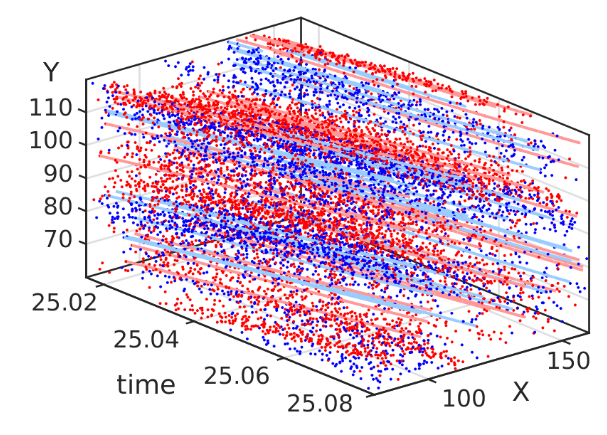}
        \caption{Events (dots) and the trajectories that they follow}
        \label{fig:gallego01}
    \end{subfigure}
    \begin{subfigure}[b]{0.48\columnwidth}
        \includegraphics[width=\textwidth]{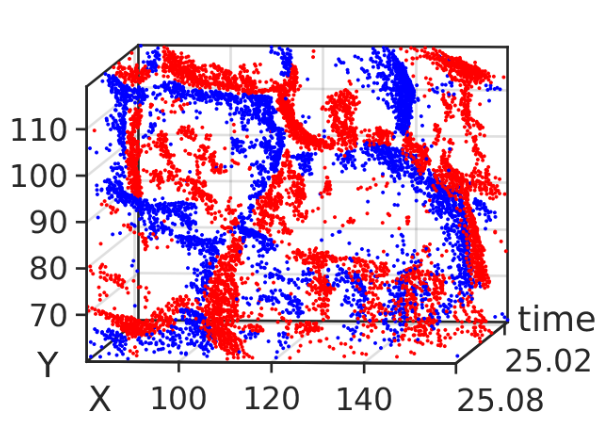}
        \caption{Events visualized along the trajectories}
        \label{fig:gallego02}
    \end{subfigure}
    \caption{Visualization of the events (positive (blue dots) and negative (red dots)) in the image plane vs. time (50 ms) \protect\cite{Gallego2017Accu}}
    \label{fig:gallego}
\end{figure}

In \cite{Gallego2017Accu} the angular velocity of an event camera was estimated based on contrast maximisation by transforming events along their trajectories on the time-image plane space. Essentially, the authors realised that if the events were viewed along their trajectories, they would produce a crisp image in which the polarities of lined-up events would sum to maximise the contrast (Fig \ref{fig:gallego}). That paper focused on estimating angular velocity by this method under the assumption that during a camera rotation the majority of events would conform to trajectories represented by one rotation transformation. Our work follows a similar thought, except that instead of estimating camera angular velocity we seek to estimate image plane velocities. The difficulty here, of course, is that we cannot assume that all event trajectories will conform to one particular transformation.\par
Finally, we would like to mention the work of Mueggler \textit{et al.} \cite{mueggler2015lifetime}, in which a solution to the event buffering for the generation of image gradients was proposed. In that work the optical flow, calculated by locally fitting a plane to the \textit{Surface of Active Events}, was used to estimate the lifetime of events. A small part of our work involves estimating the lifetime of events via the current optical flow velocity estimate, whereby we adopt a similar approach to that proposed by Mueggler \textit{et al.}
%\FloatBarrier
%Talk about how the algorithm essentially exploits the fact that all the events generated by an object moving across the image plane have the same optical flow and that that optical flow can be represented by a plane whose normal is the flow vector. If the events generated are thus projected onto the plane, they should all accumulate along their trajectories.
%Mention the assumption of linear velocities (ie no acceleration) and how the fast update time of the algorithm approximates this)
%Maybe talk about the problem of thin slices?...
\section{Methodology}
\subsection{Structure Detection}
\label{structureDetection}
\begin{figure}
    \centering
    \begin{subfigure}[b]{1\columnwidth}
        \includegraphics[width=\textwidth]{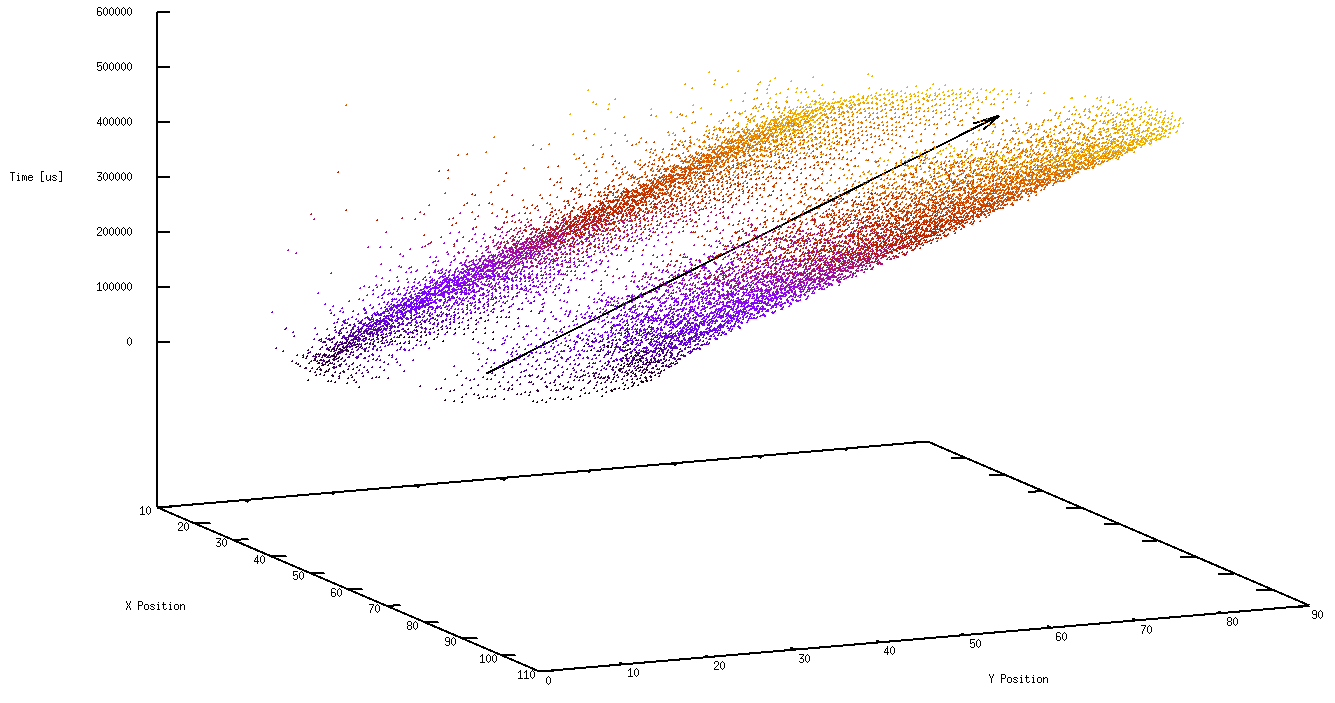}
        \caption{Events form a cylinder in space and time. The principle axis is shown as a black arrow}
        \label{fig:Event_cylinder01}
    \end{subfigure}
    \begin{subfigure}[b]{1\columnwidth}
        \includegraphics[width=\textwidth]{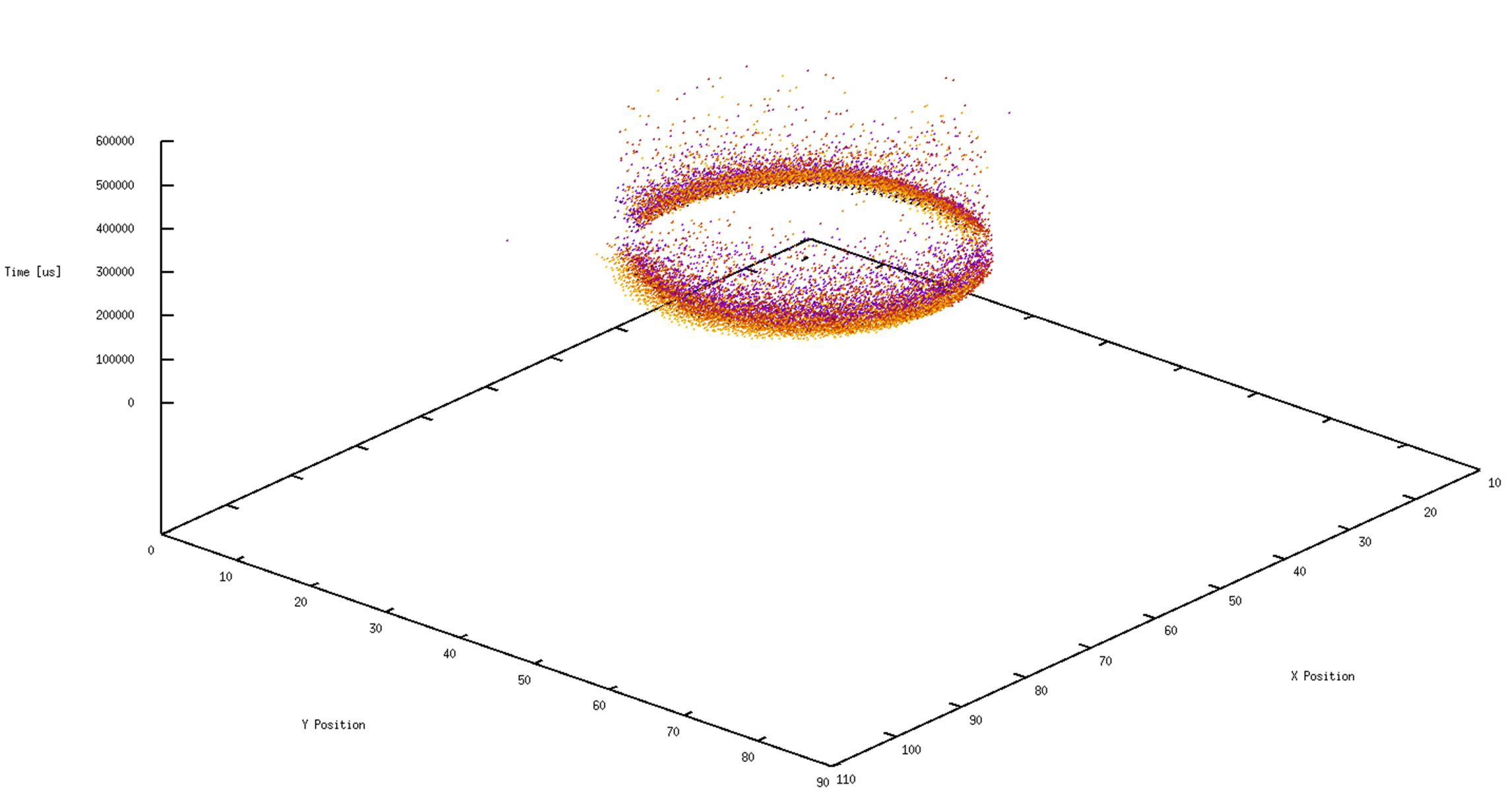}
        \caption{Events as viewed along the principle axis or along visual flow vector}
        \label{fig:Event_cylinder02}
    \end{subfigure}
    \caption{Visualisation of the events generated by a rolling ball (coloured by timestamp}
    \label{fig:Event_cylinder}
\end{figure}

\begin{figure}
	\centering
    \includegraphics[width=0.8\columnwidth]{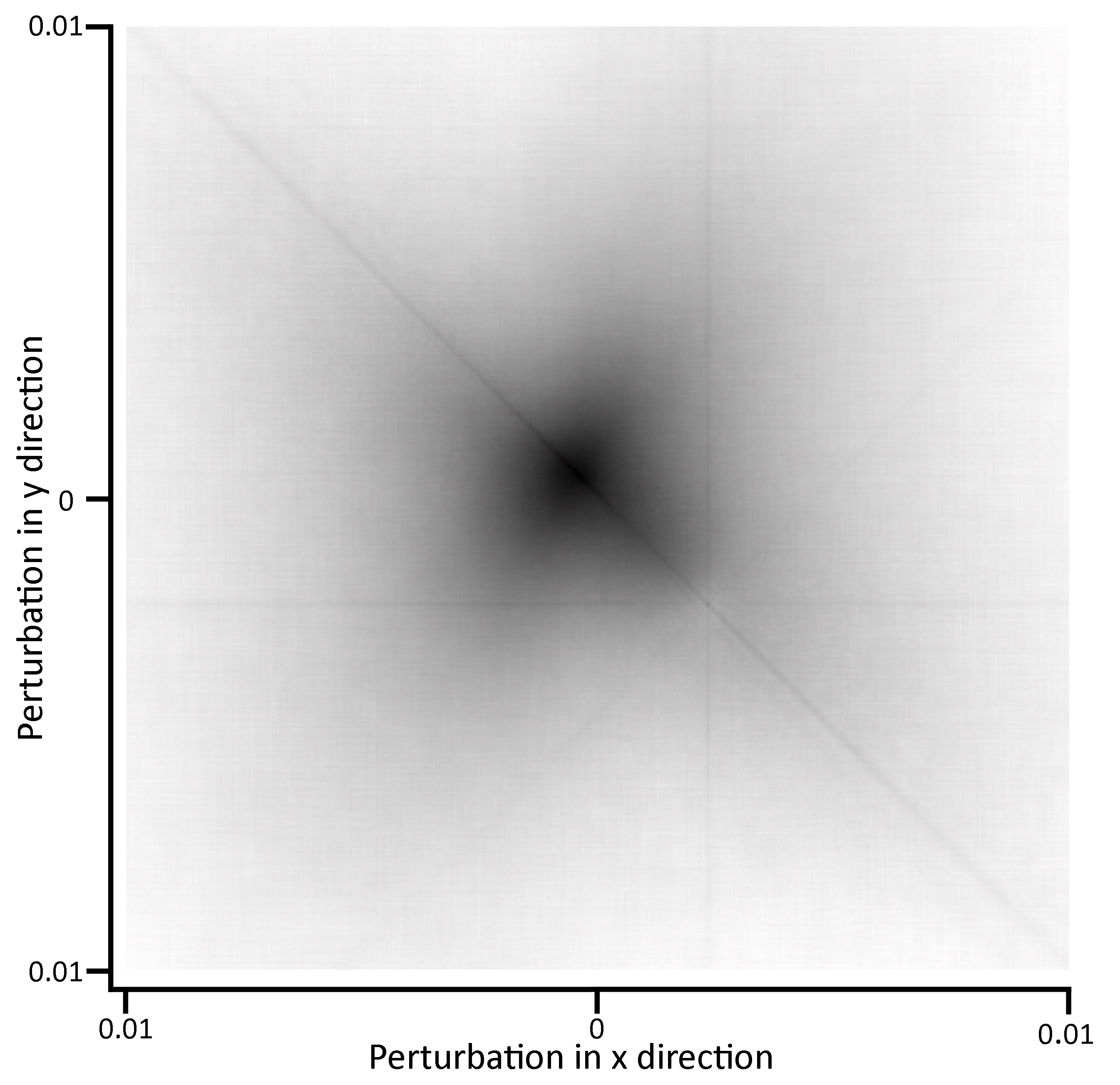}
     \caption{A map of the metrics $m$ computed across a 400x400 array of projections, whose normal $\vec{n}$ is perturbed in the x axis and y axis respectively. The events projected are from the data in Fig. \ref{fig:Event_cylinder}. Darker colors indicate a higher metric, the lines visible are due to discretisation of events. The darkest spot represents the closest projection to the true flow vector}
    \label{fig:areamap}
\end{figure}

Events generated by a DVS are represented by a vector and a sign $e=(u,v,t)^T, s$ in which $u,v$ represent the event's position on the image plane, $t$ represents the event timestamp and $s$ the event polarity (+1 for an increase in intensity, -1 for a decrease). Suppose an object generates events on it's visual structures, such as internal and external edges as it moves. Let the set of the $x,y$ coordinates of the events generated by the object's progression on the image plane by one pixel be called the 2D cross-section $C$ (Fig. \ref{fig:TrackPlanes01}). Supposing the object moves at a velocity $\vec{v}=(\vec{v}_u, \vec{v}_v)^T$ across the image plane. Then at time $t$ an event $e_t=(u_t,v_t,t_t)^T$ can be generated where 
\begin{align}
u_t=u_c+\vec{v}_ut \\
v_t=v_c+\vec{v}_vt \\
u_c, v_c \in C
\end{align}
This set of events forms a point cloud which resembles an extrusion of the original contour $C$. Thus this set of events in 3D pixel/time space will be referred to as the Event Contour Extrusion (ECE) or $E=\lbrace e_0,e_1...e_k \rbrace$ (Fig. \ref{fig:Event_cylinder}). The principle axis of this extrusion runs parallel to the parametrically defined line $l$
\begin{align}
u&=\vec{v}_ut\\
v&=\vec{v}_ut\\
t&=t
\end{align}
Therefore $l=(u,v,t)^T=t(\vec{v}_u, \vec{v}_v, 1)^T=t\vec{n}$. By projecting the events in $E$ along $t\vec{n}$ the time dimension can thus be collapsed for the ECE generated by a structure moving with constant velocity with respect to the DVS, revealing the original contour $C$. This projection is defined as
\begin{align}
proj_{\vec{n}}(e)&=(u-\vec{v}_ut, v-\vec{v}_vt, 0)\\
&=\begin{bmatrix}
    1	&0	&-\vec{v}_u\\
    0	&1	&-\vec{v}_v\\
    0	&0	&0
\end{bmatrix}
\begin{bmatrix}
u\\
v\\
t
\end{bmatrix}
\end{align}
If the events are now projected onto a horizontal plane spanning $u,v$ using $proj_{\vec{n}}(e)$, events generated by the same feature of the object will project onto the same parts of the plane. Summing these accumulated events gives us 
\begin{align}
f(x,y)=\sum\limits_{i=0}^{k}
\begin{cases} 
      0 & proj_{\vec{n}}(e_k) \neq \vec{q} \\
      s_k & proj_{\vec{n}}(e_k) = \vec{q} \\
\end{cases}
\end{align}
for every location $\vec{q}=(x,y)^T$ on the plane. Summing the squares of these sums gives us a metric 
\begin{align}
m=\sum\limits_{i=-\infty}^{+\infty}\sum\limits_{j=-\infty}^{+\infty} f(i,j)^2
\end{align}
If one now considers the set of $n$ projections $F_p=\lbrace f_{p1}...f_{pn} \rbrace$ slightly perturbed relative to the true  projection $f_{p0}$, it becomes apparent that the corresponding metric $m_{p0}...m_{pn}$ will change as the accumulated sums change. Further, as $k$ grows large, the metric of the most representative projection $m_0$ will become greater than that of any of the other projections $m_{p1}...m_{pn}$. This is obvious, since any projection which is inclined to the principle axis will eventually begin to have the projected events drift away, in the equivalent phenomenon of motion blur \cite{Gallego2017Accu}, while the true projection will continue to gather accumulations. Thus, finding the projection with the maximum value of $m$ given sufficient events is equivalent to finding the optic flow of a given structure moving across the image plane (Fig. \ref{fig:areamap}).\par
\begin{figure}
	\centering
    \includegraphics[width=0.8\columnwidth]{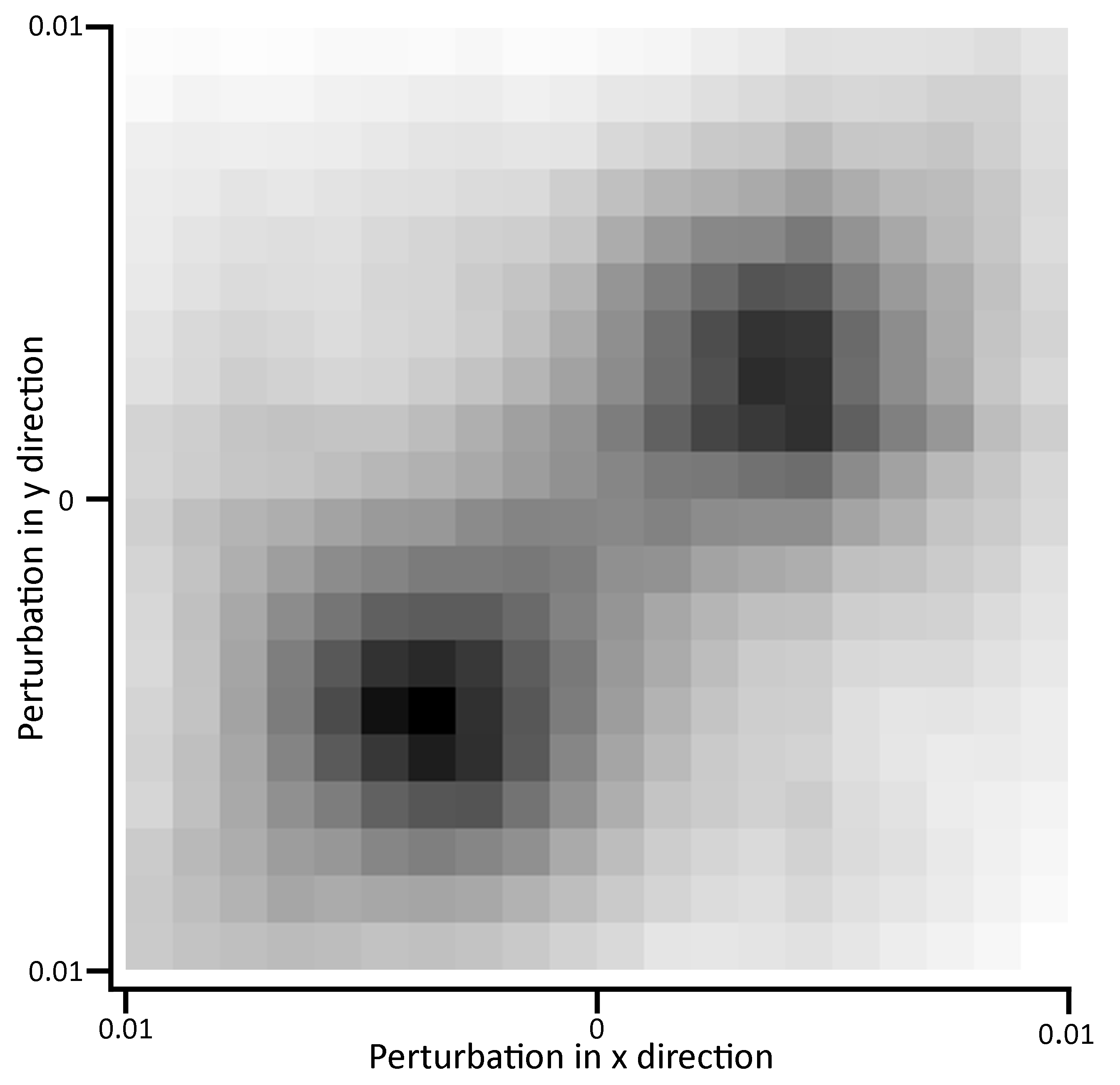}
     \caption{A map of the metrics $m$ computed across a 20x20 array of projections, whose normal $\vec{n}$ is perturbed in the x axis and y axis respectively. The events projected were generated by two structures with opposite velocities moving across the image plane - this is clearly visible in the two maxima. Darker colors indicate a higher metric. The darkest spots represent the closest projections to the true flow vectors}
    \label{fig:areaMap2}
\end{figure}

In fact, this property remains even with multiple structures with different velocities. In practice, it requires more events for the best projections to become apparent, however given a sufficient number of events the number of local maxima in the set of projections will match the number of structures with a distinct optic flow velocity and those maxima will correspond to the best approximations to the optic flow of each structure respectively. This can be seen clearly in Fig. \ref{fig:areaMap2}.
\subsection{Structure Tracking}
Clearly, simple projections of events as defined above can only work under the assumption that the structures generating events are travelling with a constant velocity. A more complicated motion model that allows for acceleration could be developed, by adding a further dimension to the projections. Further it is possible for the contours of structures in the scene to change, rendering previous accumulations of events a liability. Therefore this concept of projection metrics is only used in the initialisation stage of our algorithm. Since both the velocity and the contours of structures generating events can be assumed constant over short periods of time, we introduce an iteratively improving tracking mechanism. The method described in \ref{structureDetection} provides a good estimate of the initial flow vector as well as an outline of the edges that generated the events. This outline together with the corresponding projection vector is then used to predict the location of incoming events. The accuracy of this prediction is then used to both update the inclination of the projection as well as the outline of the structure.

%\FloatBarrier
\section{Algorithm and Implementation}
\label{sec:Algorithm}
\begin{figure}
	\centering
    \includegraphics[width=\columnwidth]{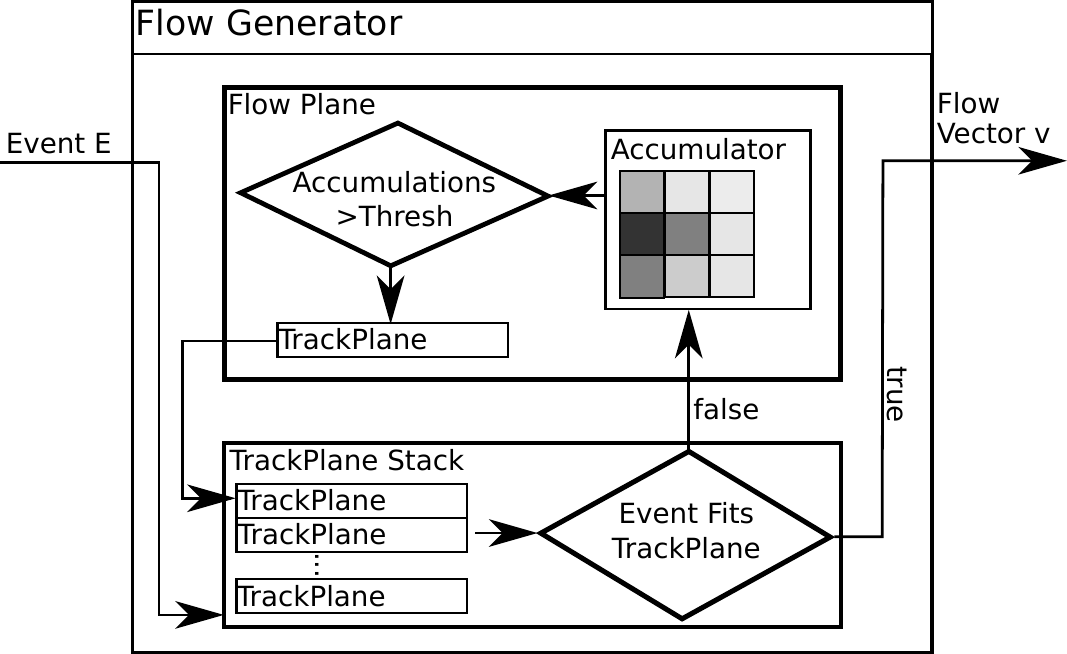}
     \caption{Structure of the algorithm. Incoming events are matched to Track Planes and failing a match added to the initialisation or Flow Plane module. On being allocated to a plane they are given the corresponding flow vector.}
    \label{fig:Algorithm_Flow_Chart01}
\end{figure}

The algorithm for optical flow detection is split into two separate modules (Fig. \ref{fig:Algorithm_Flow_Chart01}) as described in section 3. As events are added they are first matched against existing structures or \textit{Track Planes} in the Track Plane stack. If matching fails, it is assumed that this event belongs to a previously undiscovered structure moving across the image plane and is added to the \textit{Flow Plane} module, which seeks to discover structures with a particular flow velocity in the scene and generate new Track Planes accordingly. When an event is allocated a Track Plane, it is allocated the corresponding flow vector estimate of that Track Plane.\par
In some cases, the same structure will be segmented into several distinct Track Planes (for example in the case of an object entering the frame). Thus the algorithm also has a mechanism to merge several Track Planes belonging to the same structure. Track Planes that receive less than $n\%$ of the events expected (based on their velocity and the number of accumulator cells) are removed from the stack.
\subsection{Flow Plane}
The Flow Plane module consists of an array ($M(x,y)$) of $n$x$n$ flow projections. Each of these is perturbed evenly by some factor, such that the angle between corner planes is some value $r$. (While this value is arbitrary, we typically perturb the planes such that $r=\pi$. Since any projection that is tilted at angle $\frac{\pi}{2}$ from $(0,0,1)^T$ has infinite velocity, this covers the full range of possible velocities). Incoming events are now projected to each flow plane, where accumulating values are summed, and each such sum is squared and summed, forming the metric space $M(x,y) = m, 0 \leq x,y \leq n$. The corresponding metrics thus change value with each incoming event. After each event the global maximum in $m=(x,y)^T, M(x,y)=\textrm{max}(M)$ is found. In early stages the location of this maximum changes frequently, however as more events are added it converges. Thus, after the maximum has remained constant for $p$ events, it is considered to contain a structure(s) and valid flow vector estimate.\par
\begin{figure}
	\centering
    \includegraphics[width=\columnwidth]{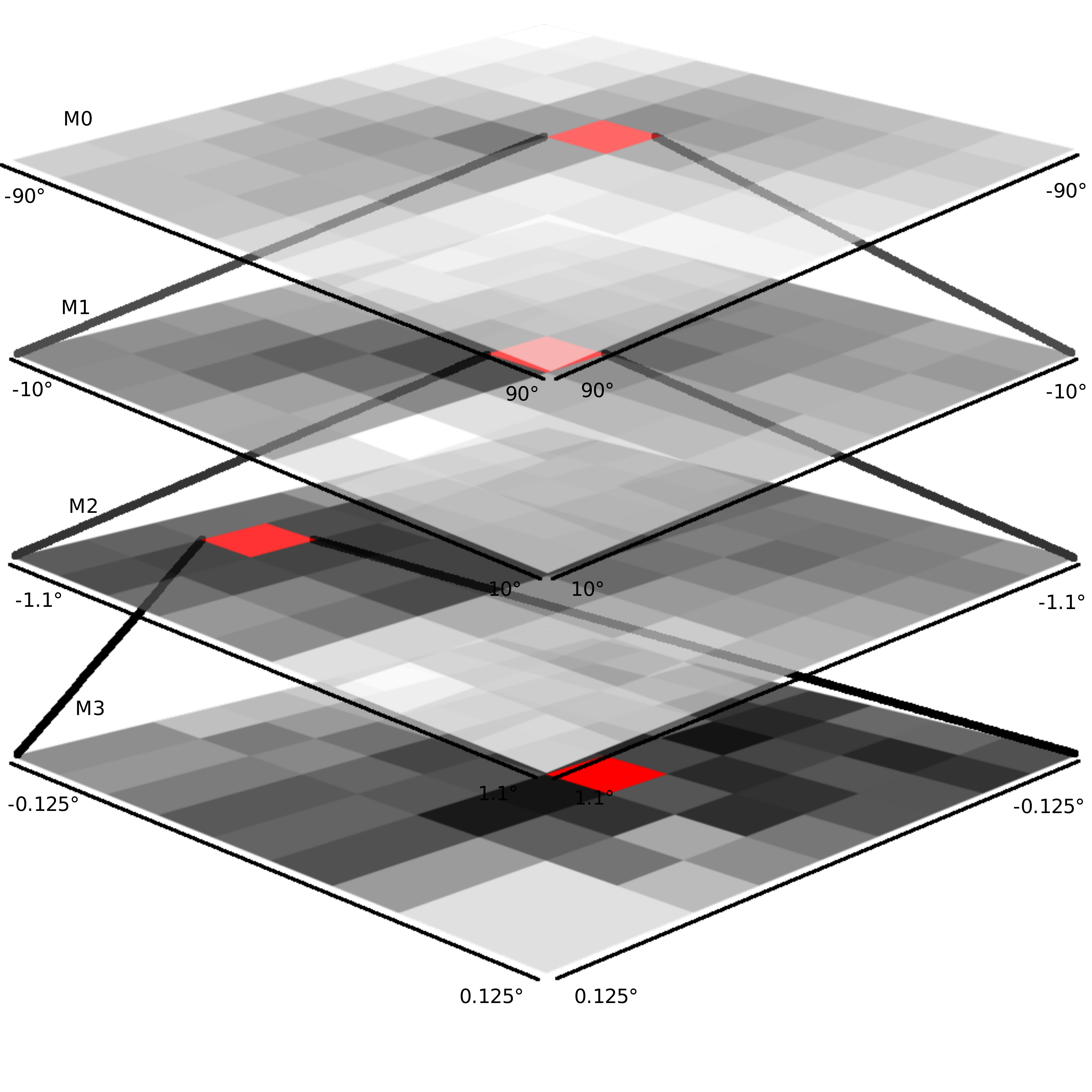}
     \caption{Stack of projection metric arrays $M(x,y)$. The top metric array $M_0$ has a low resolution ($180^\circ$), but is expanded into $M_1$ ($20^\circ$) when a best projection is found. Darker values indicate a higher value. Global maximum of each $M$ is shown in red.}
    \label{fig:Stack01}
\end{figure}

This estimate however, tends to be poor, especially if the value of $n$ is low. Therefore, in the next stage, the events that correspond well to the chosen projection are taken and separately projected by a new $n$x$n$ array of projections, with a perturbation range of $\frac{r}{q}$ for some factor $q$ (Fig. \ref{fig:Stack01}). This has two effects: for one the new estimate is much closer to the true velocity of the underlying structure since the resolution is improved, for another the new estimate is much more able to represent the structure since only events corresponding to the chosen flow projection are chosen for this refining step.\par
The way the most associated events are chosen is by inspecting $f(x,y) \forall x,y \in \mathbb{Z}$. If $f(x,y)>\mu_f+\sigma_fw$, where $\mu_f$ and $\sigma_f$ are the average value and standard deviation of the values of $f(x,y)$ respectively and $w$ is a thresholding factor, then the events accumulated at that location are taken to be part of the structure located. All events at locations connected to the locations found are also added to the set of associated events $E_{assoc}$ akin to flood-fill.\par
After several such refining cycles, $E_{assoc}$ and the located projection $\vec{n}$ are bundled into a new Track Plane, which is then added to the Track Plane stack.\par
The Flow Plane is now cleared and the set of remaining events $E_{\overline{assoc}}$ is projected onto it again. Since this set of events is no longer polluted by the structure detected in $E_{assoc}$, it is much more likely to converge on any remaining structures fast. Finally, events which are due to noise and should not be associated to any structure are eventually flushed out of the Flow Plane. This is done with a simple lifespan threshold.
\subsection{Track Plane}
\begin{figure}
	\centering
    \includegraphics[width=\columnwidth]{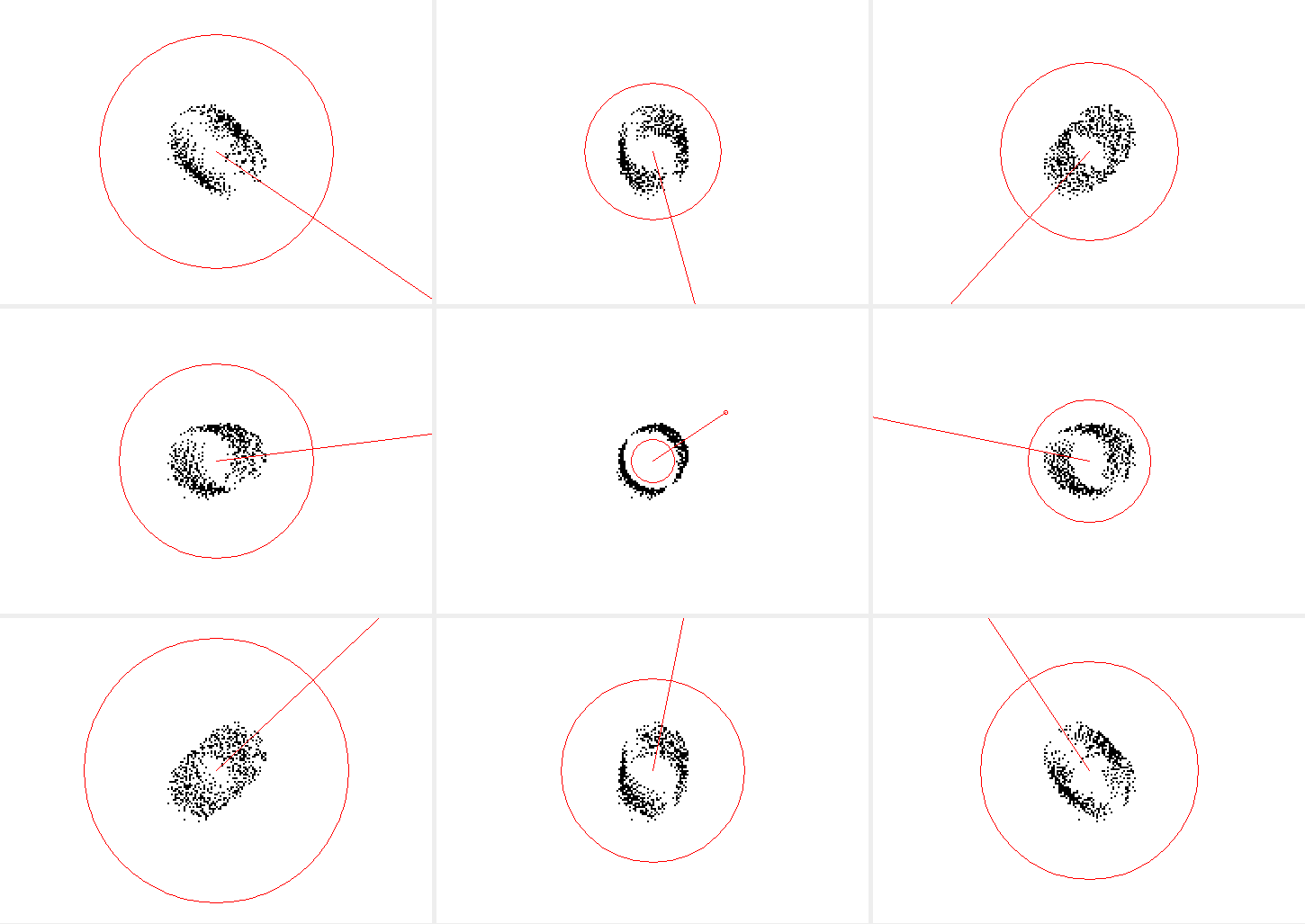}
     \caption{A 3x3 array of projections created with the data from Fig. 2. The centre projection best represents the optic flow, as seen by the crisp outline of the accumulator cells. The surrounding projections are perturbed by $h=0.02^\circ$ and are worse estimates, as shown by the ``motion blur" of viewing the cylinder from a side-on angle. The circles depict a scaled version of the magnitude and the red arrows the unscaled flow vector for each respective projection.}
    \label{fig:TrackPlanes01}
\end{figure}

Each Track Plane consists of $m^2$ projections arranged in an $m$x$m$ array (Fig.\ref{fig:TrackPlanes01}). The projection in the centre of this array represents the current best prediction of the optical flow, with the surrounding projection perturbed by a factor $h$. These all project the current set of associated events and those coordinates $(x,y)$ to which they are transformed become the set of accumulator cells $A=\lbrace(x_0,y_0)^T,(x_1,y_1)^T\cdots(x_n,y_n)^T\rbrace$.\par
As events enter the algorithm (Fig. \ref{fig:Algorithm_Flow_Chart01}), they are matched against existing Track Planes. This happens by projecting the given event by the current best estimate. If the projection returns a location $(u,v)^T$ in $A$ the event is considered to be associated with the plane and accordingly added to $A$. If it misses, it raises the value of the accumulator at $(u,v)$. This way, if the value of the accumulator at $(u,v)$ passes a threshold it can be considered to be a valid accumulator cell and thus the contour can evolve.\par
A projection in the array is considered to be the best projection if the number of hits surpasses a given percentage of the number of accumulator cells ($10\%$ typically works well). When this occurs, the projections are recalculated, centred around the new best estimate and the accumulators are regenerated based on the set of associated events from the new best projections accumulator. If the best projection is the previous best projection, the perturbation factor is halved and conversely if the new best projection was one of the surrounding projections the perturbation factor is doubled. This way the Track Plane is able to iteratively improve the estimate if the current estimate is good and conversely improve the rate of exploration if it is bad (generally because of acceleration).\par
Clearly, the Track Plane would be highly sensitive to changes in the velocity of the tracked structure if it kept the events in $A$ indefinitely. To this purpose a lifetime for events is defined, after which events are removed from $A$. Similarly to \cite{mueggler2015lifetime} this lifetime is calculated by taking the current flow velocity and calculating how long it would take for the structure to cross over $p$ pixels (we have found $p=3$ to work well). Thus if the structure is moving very slowly or even standing still, the lifetime is very long and vice versa.
%\FloatBarrier
\section{Results}
\begin{figure*}[h]
        \centering
        \begin{tabular}{cM{35mm}M{35mm}M{35mm}M{35mm}}
           \toprule
           Shape & \multicolumn{2}{c}{Magnitude} & \multicolumn{2}{c}{Angle}\\
            &SOFAS (this paper)&Lucas-Kanade&SOFAS(this paper)& Lucas-Kanade \\
            \midrule
            \includegraphics[width=1cm]{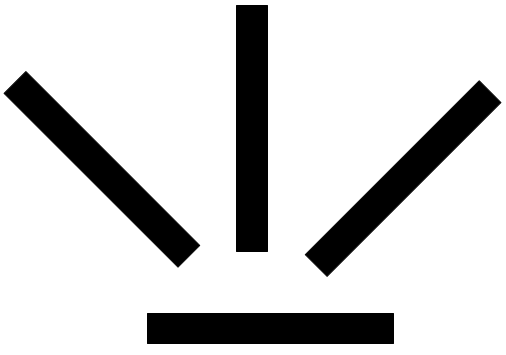} & \includegraphics[width=4cm]{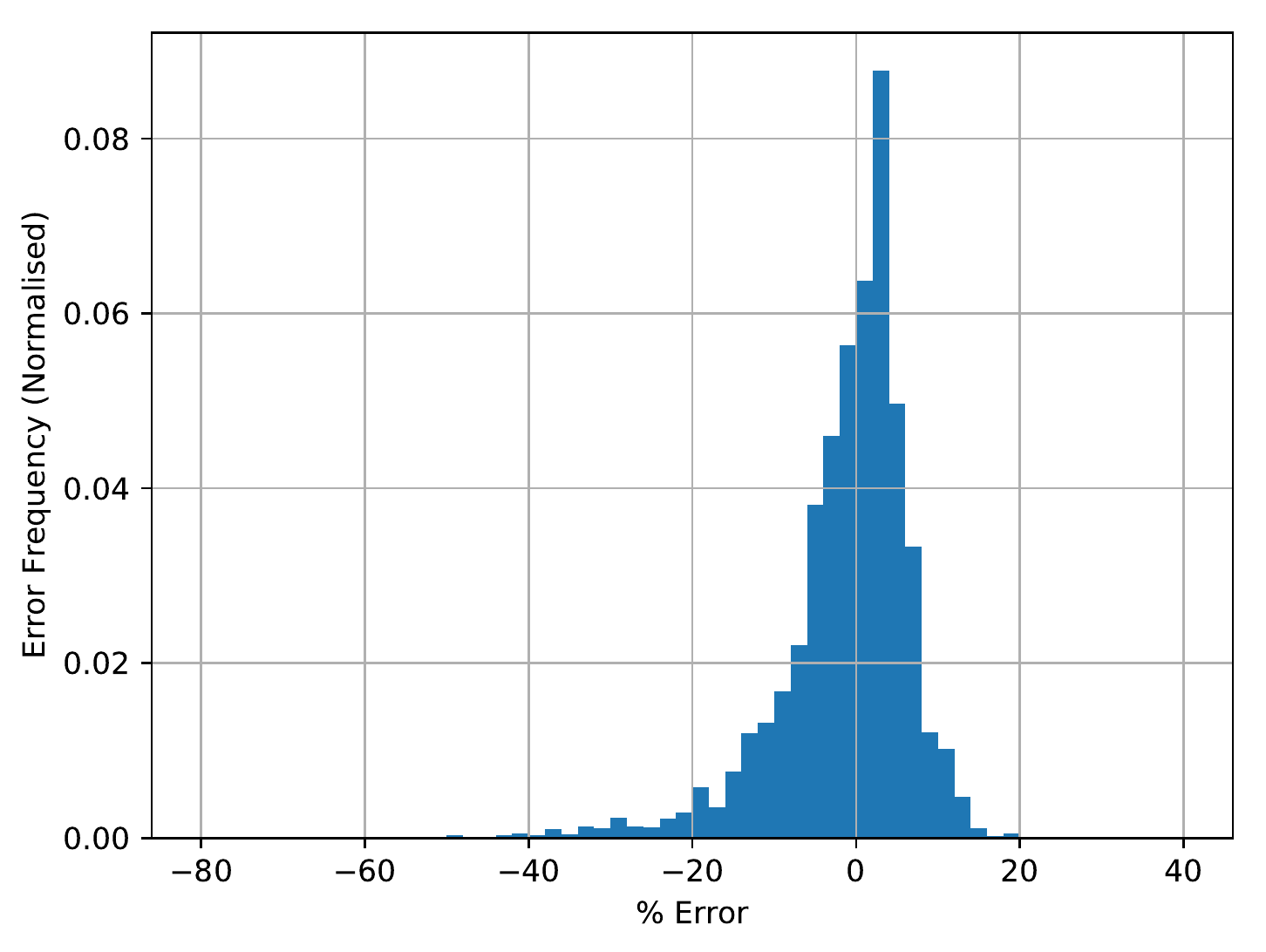} & \includegraphics[width=4cm]{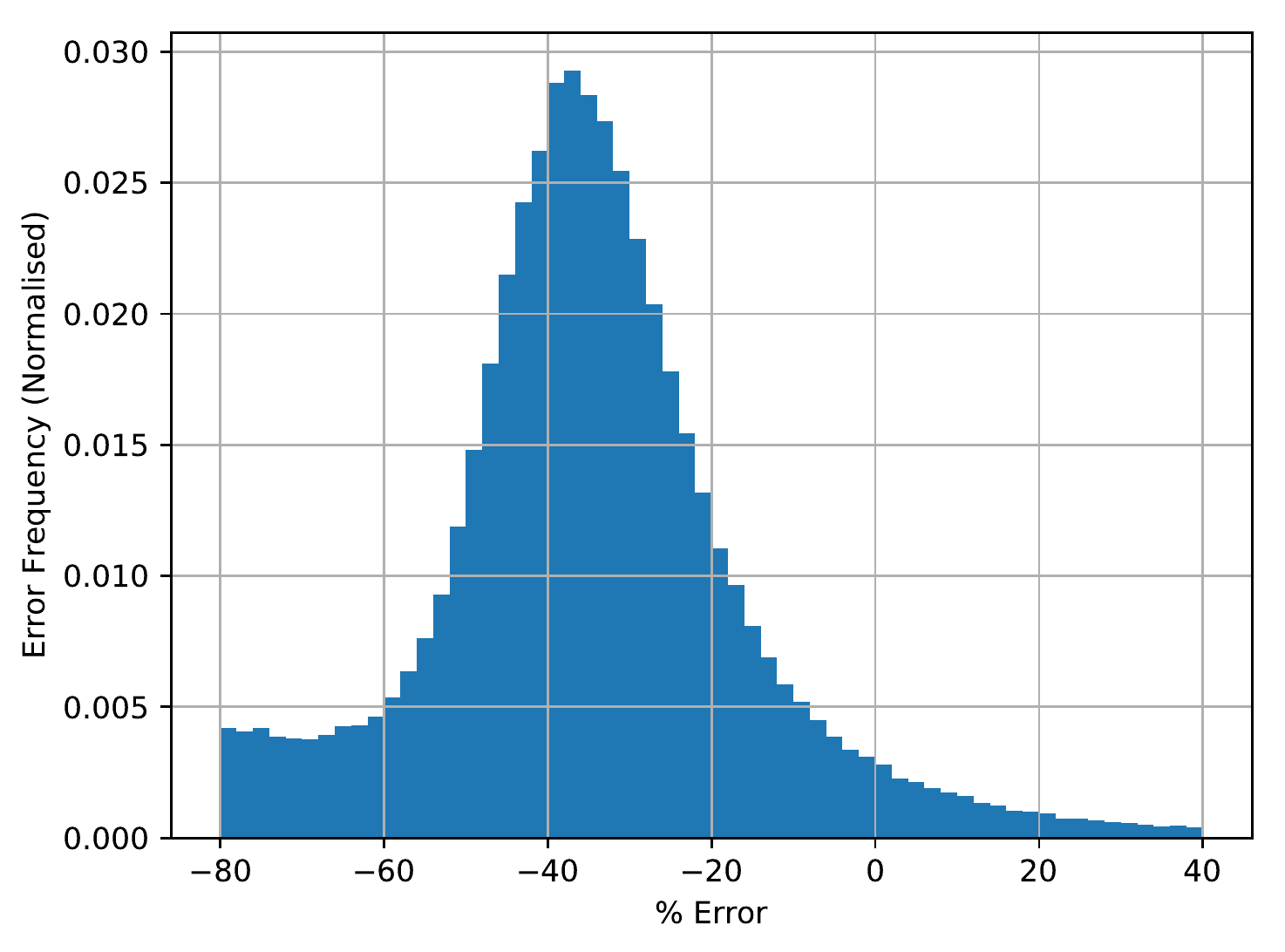} & \includegraphics[width=4cm]{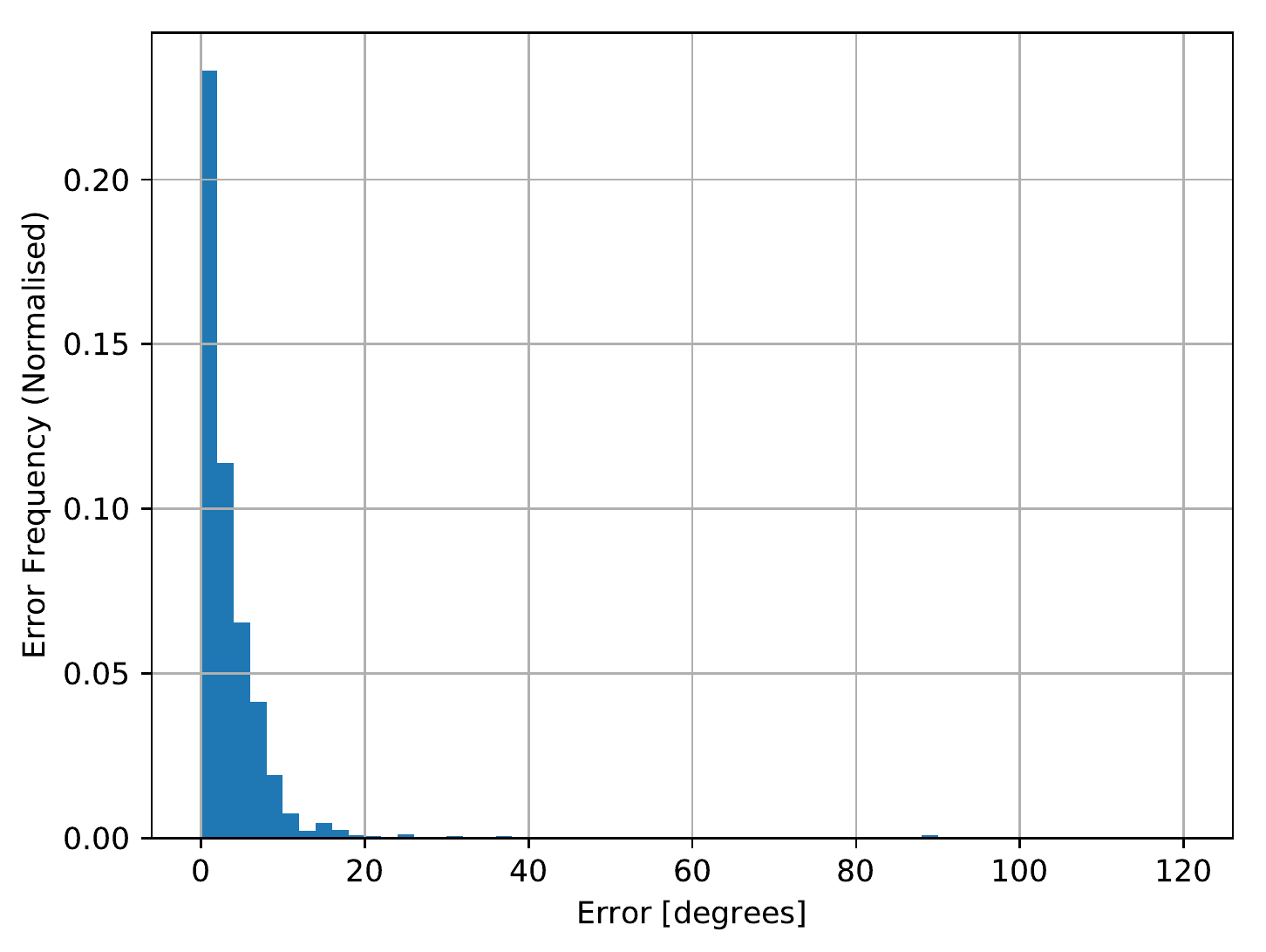} &  \includegraphics[width=4cm]{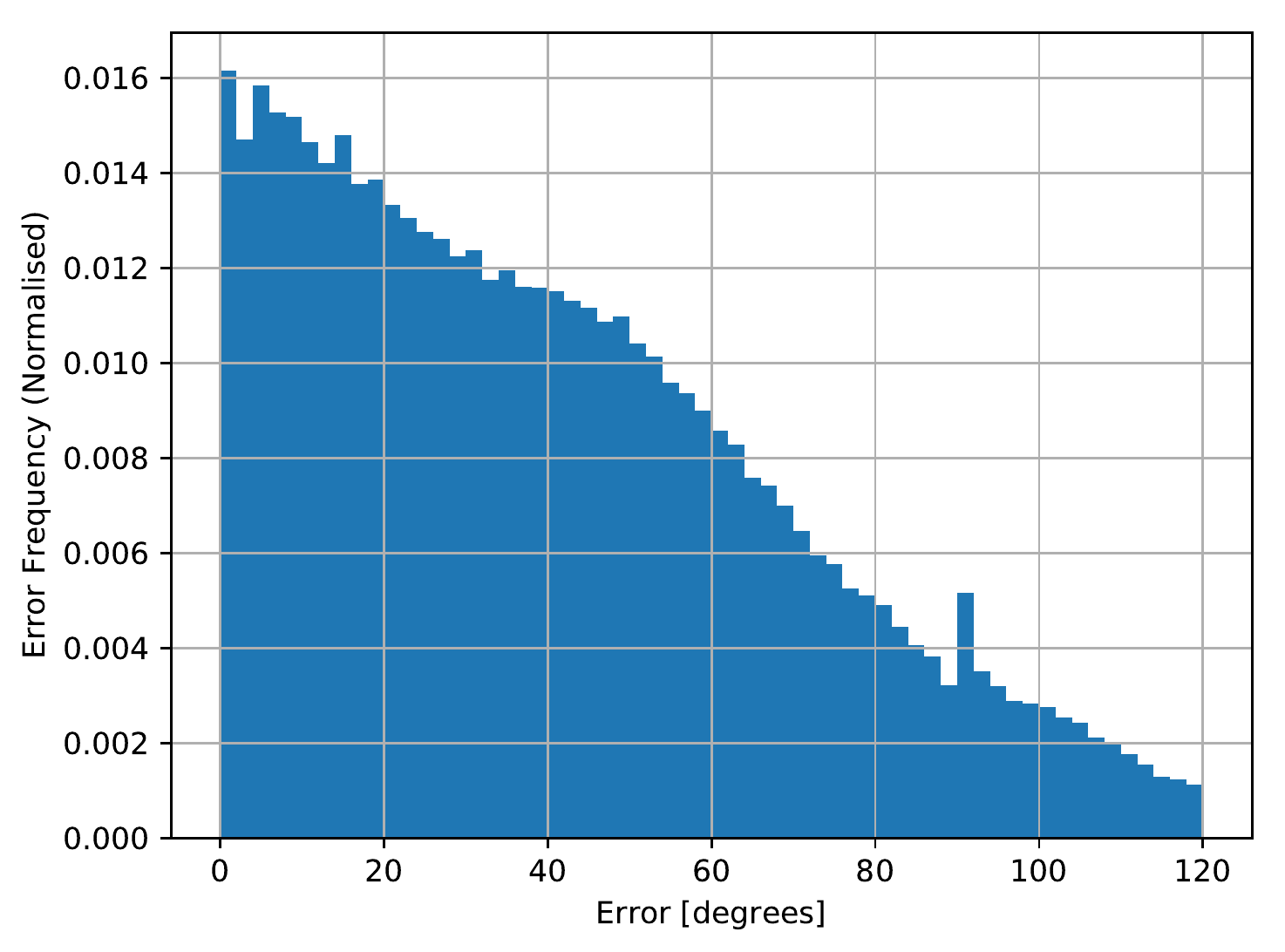} \\
            \includegraphics[width=1cm]{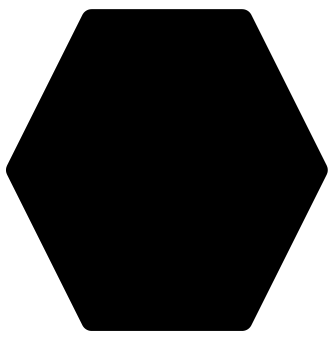} & \includegraphics[width=4cm]{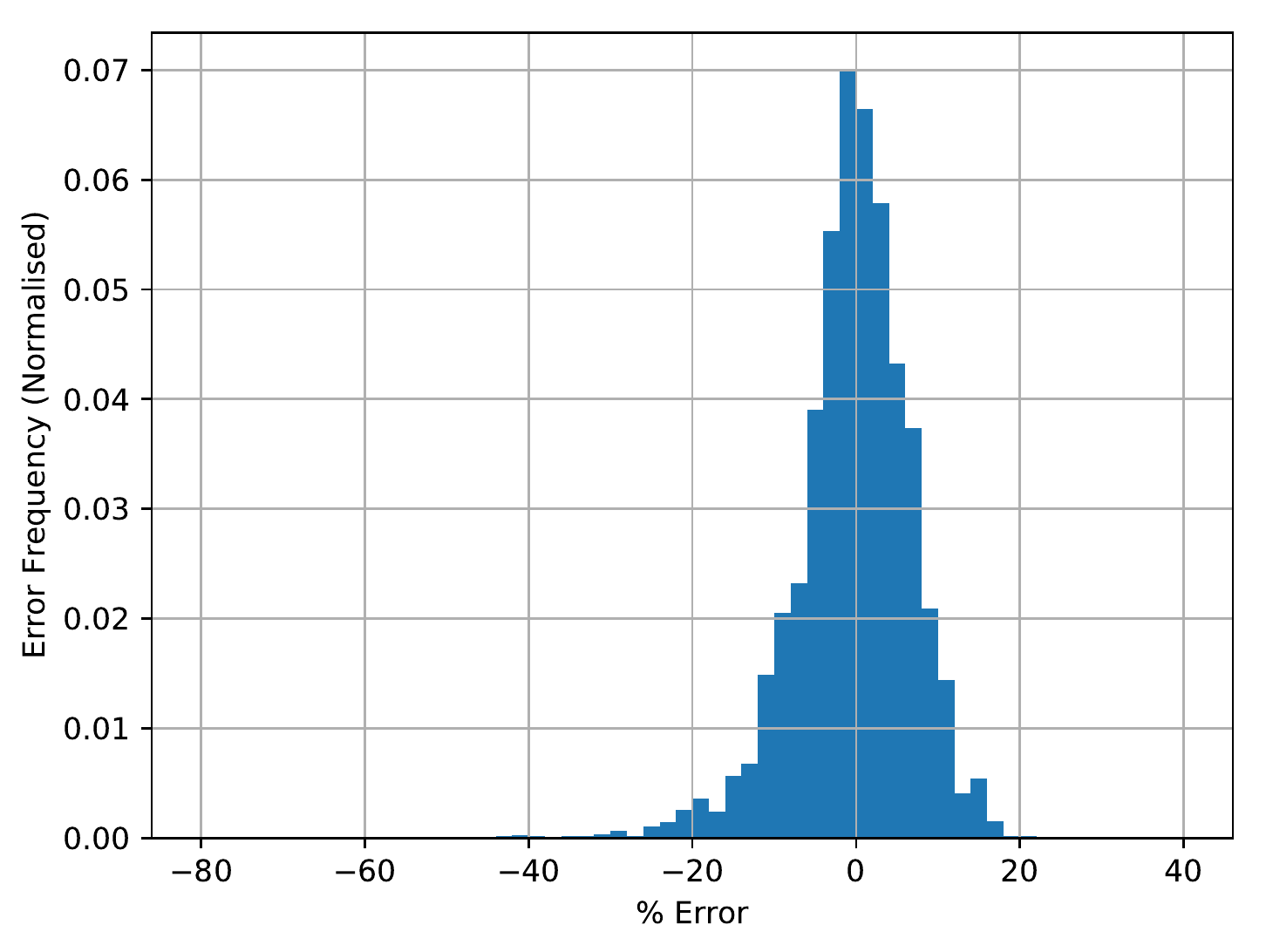} & \includegraphics[width=4cm]{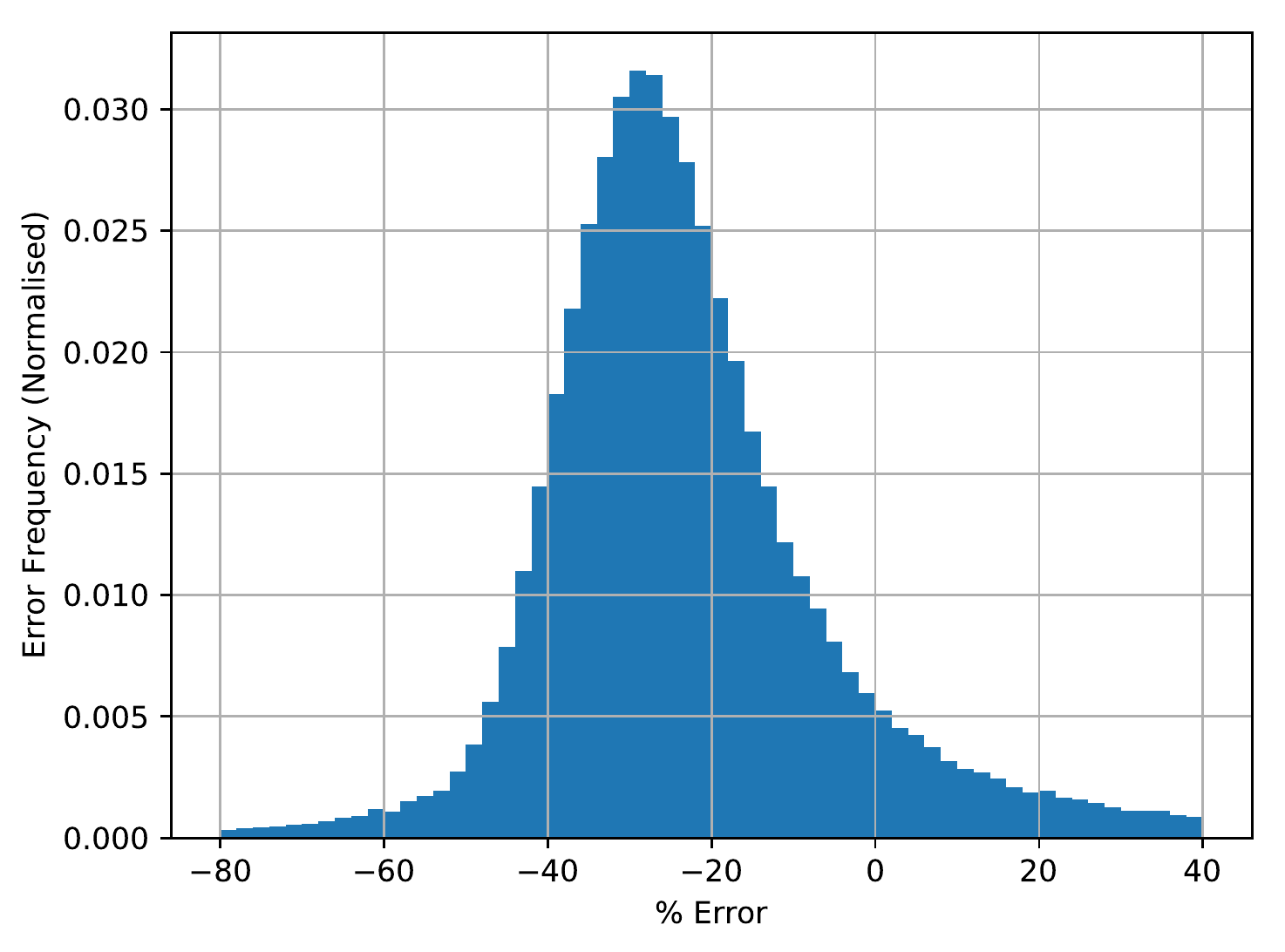} & \includegraphics[width=4cm]{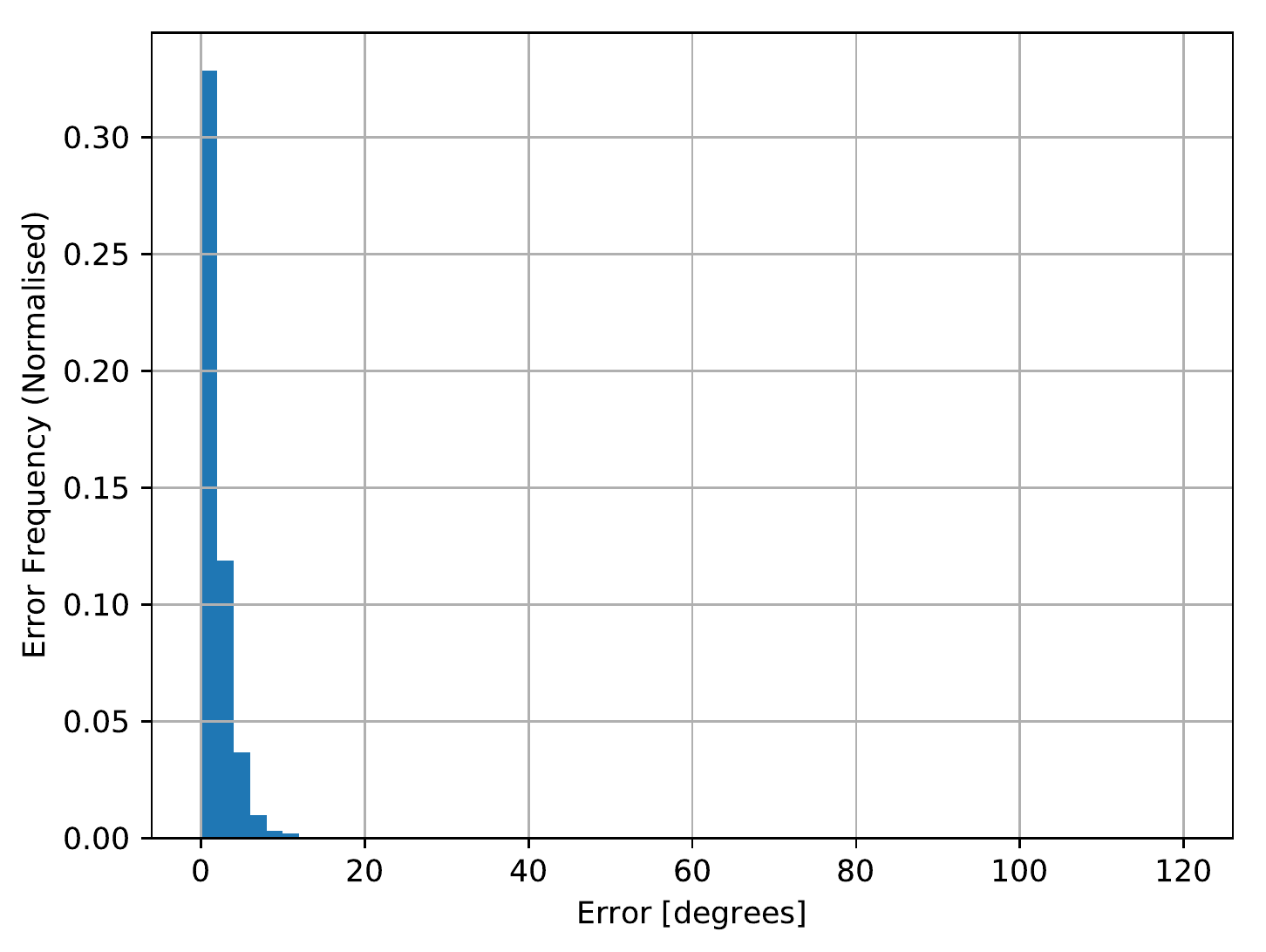} & \includegraphics[width=4cm]{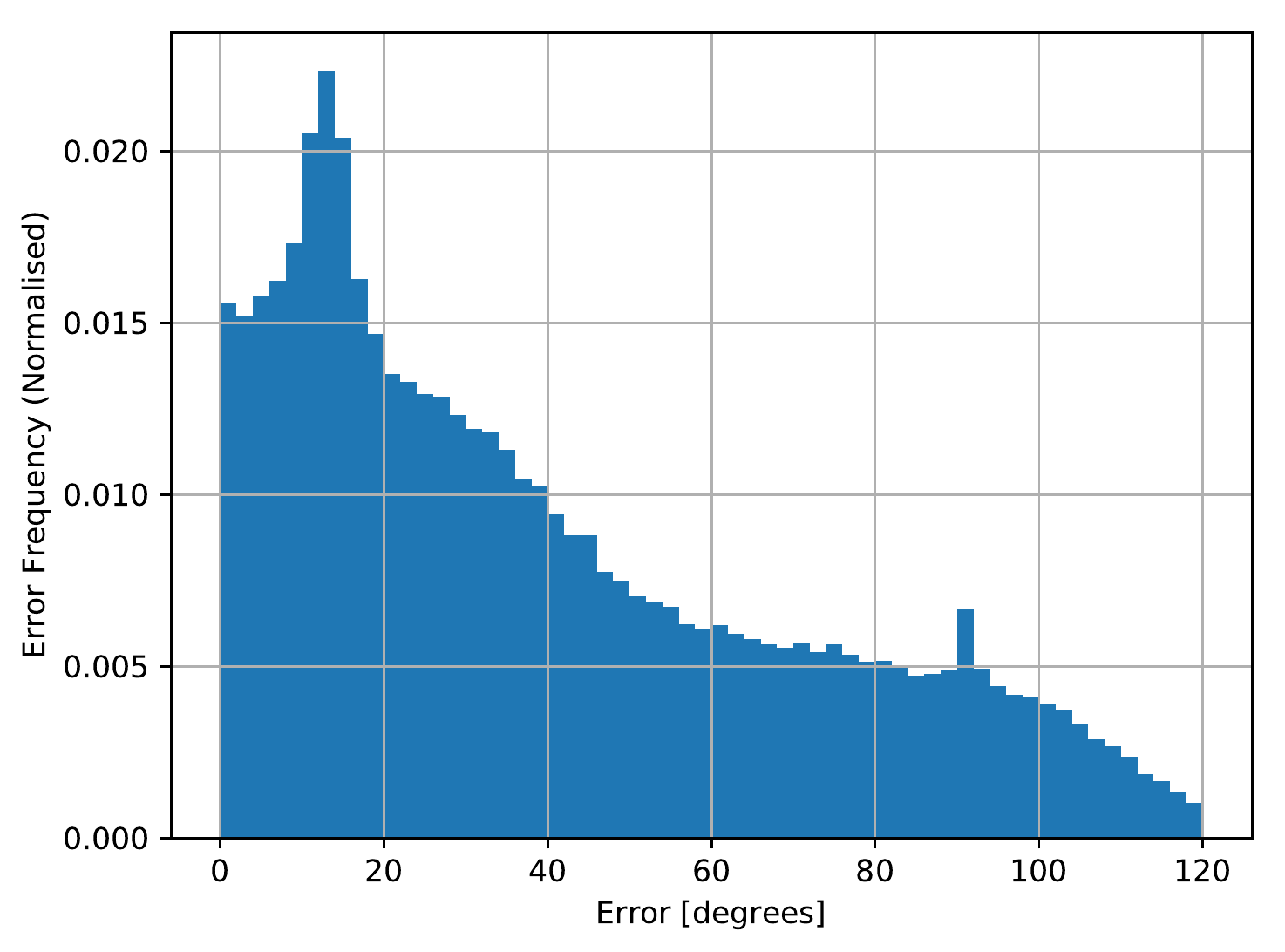} \\
            \includegraphics[width=1cm]{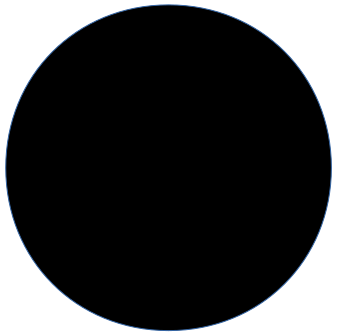} & \includegraphics[width=4cm]{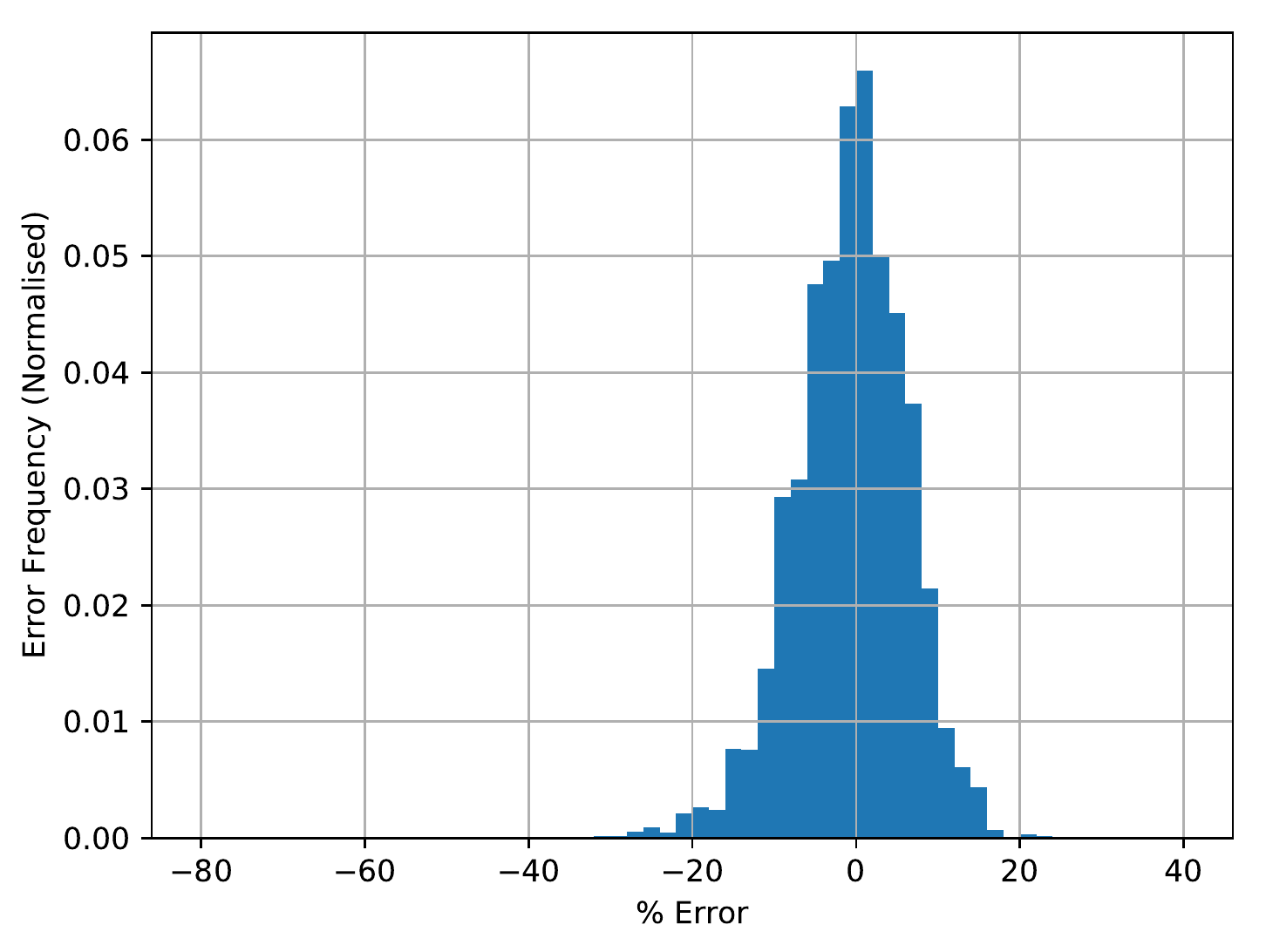}  & \includegraphics[width=4cm]{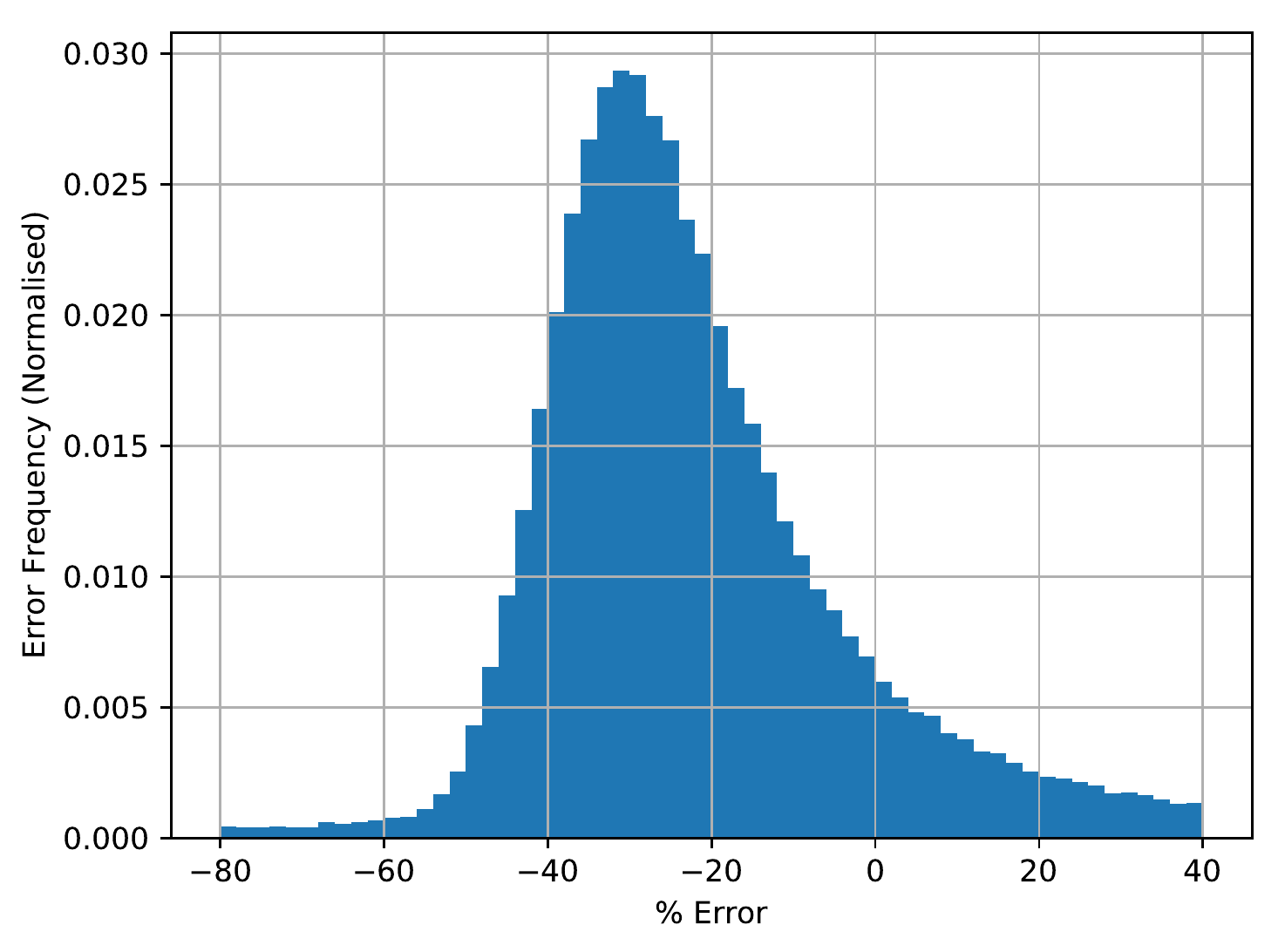} & \includegraphics[width=4cm]{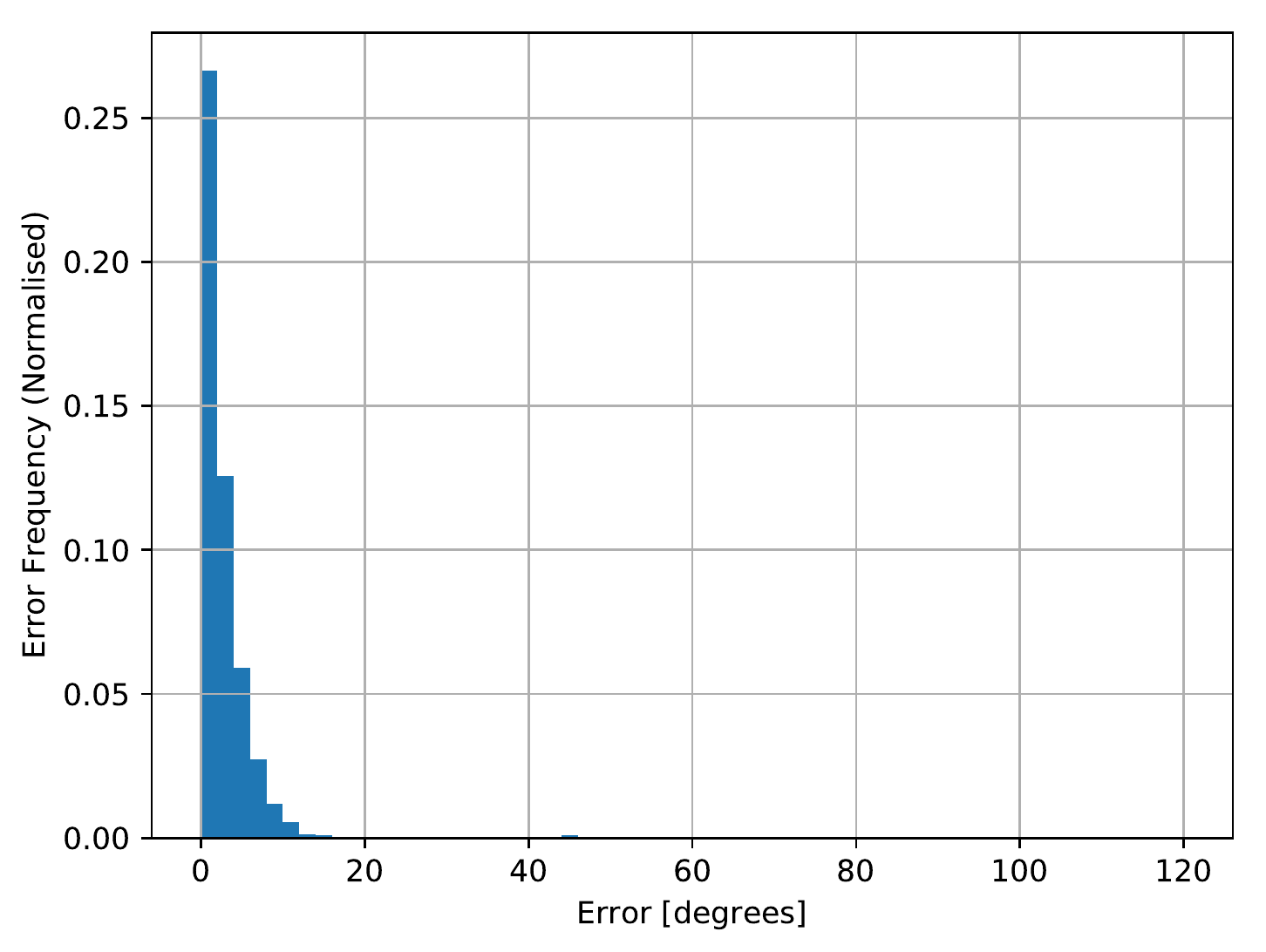} & \includegraphics[width=4cm]{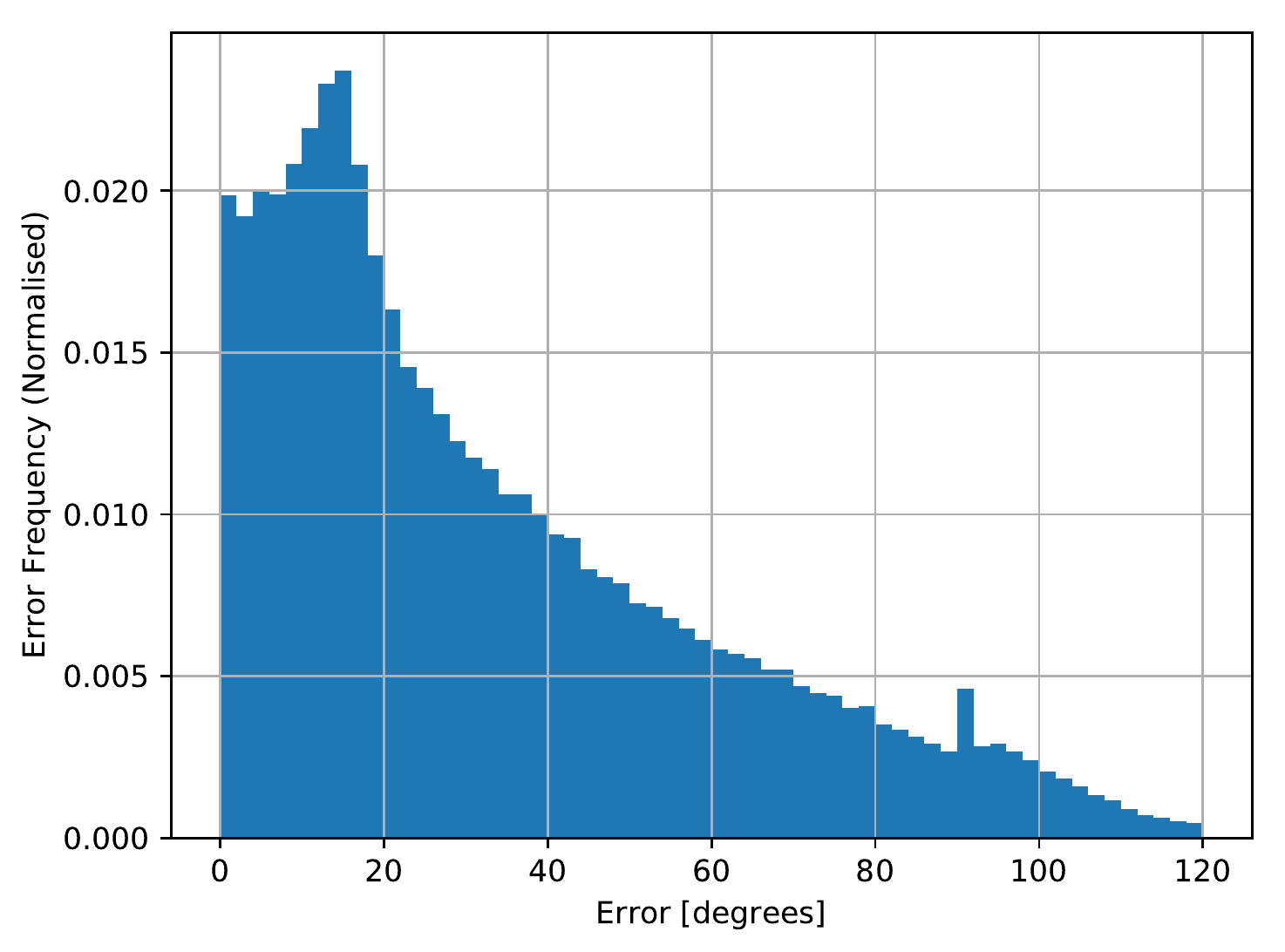} \\
            \includegraphics[width=1cm]{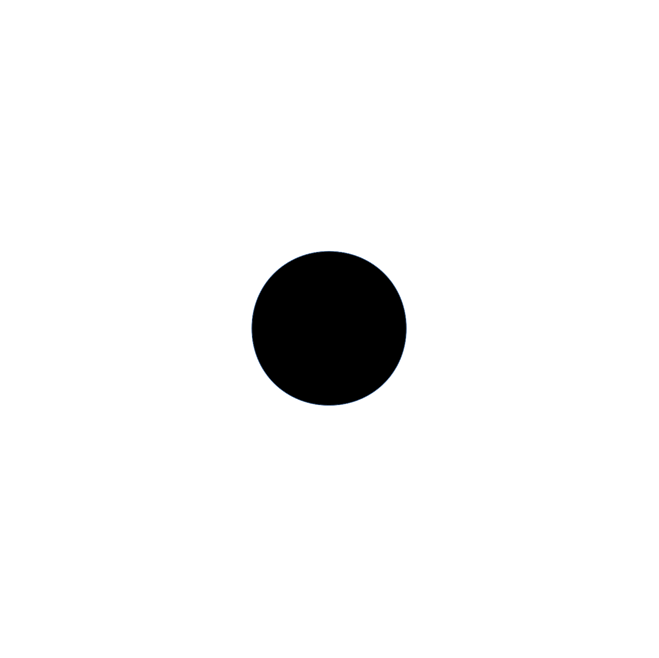} & \includegraphics[width=4cm]{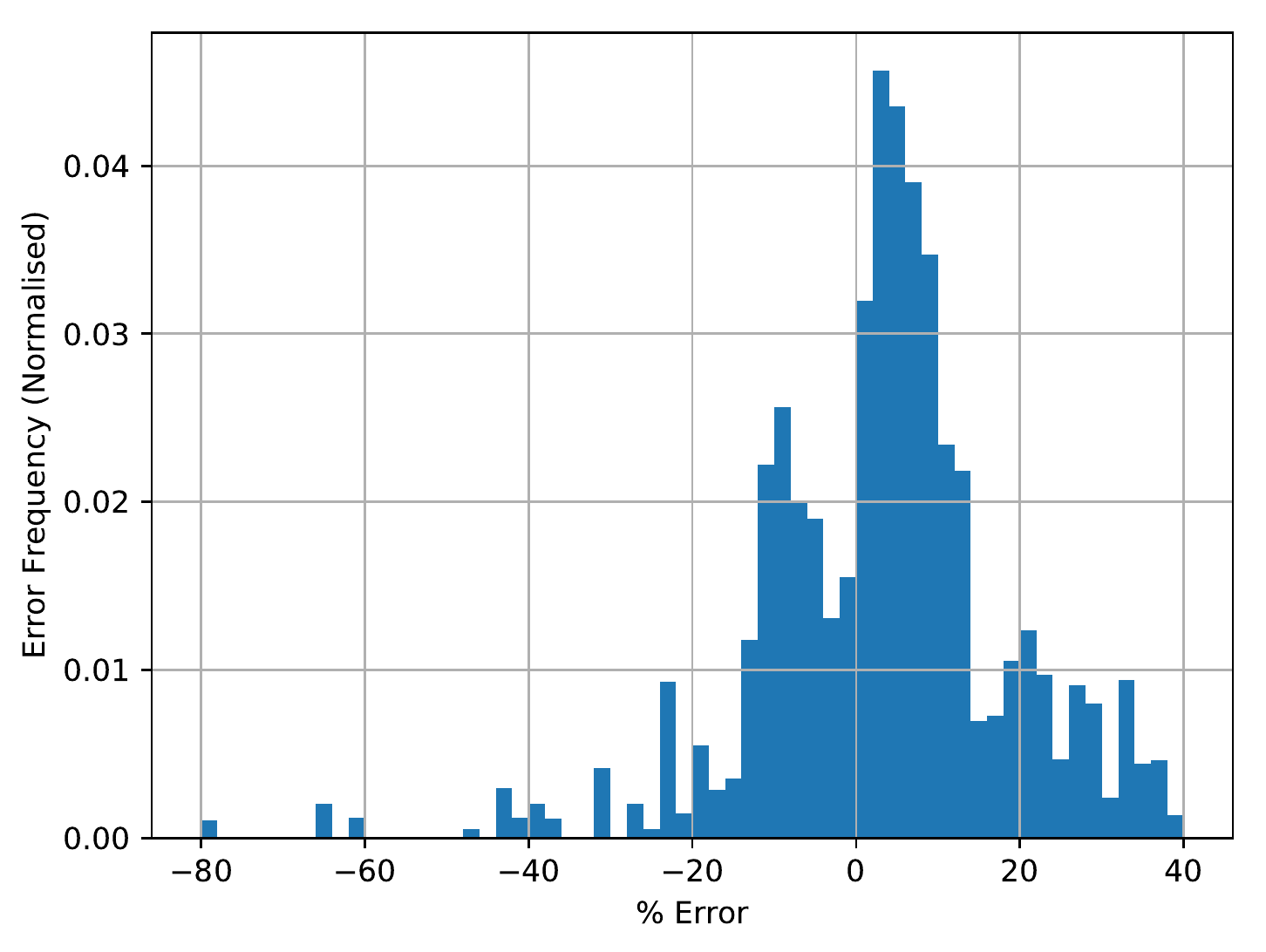} & \includegraphics[width=4cm]{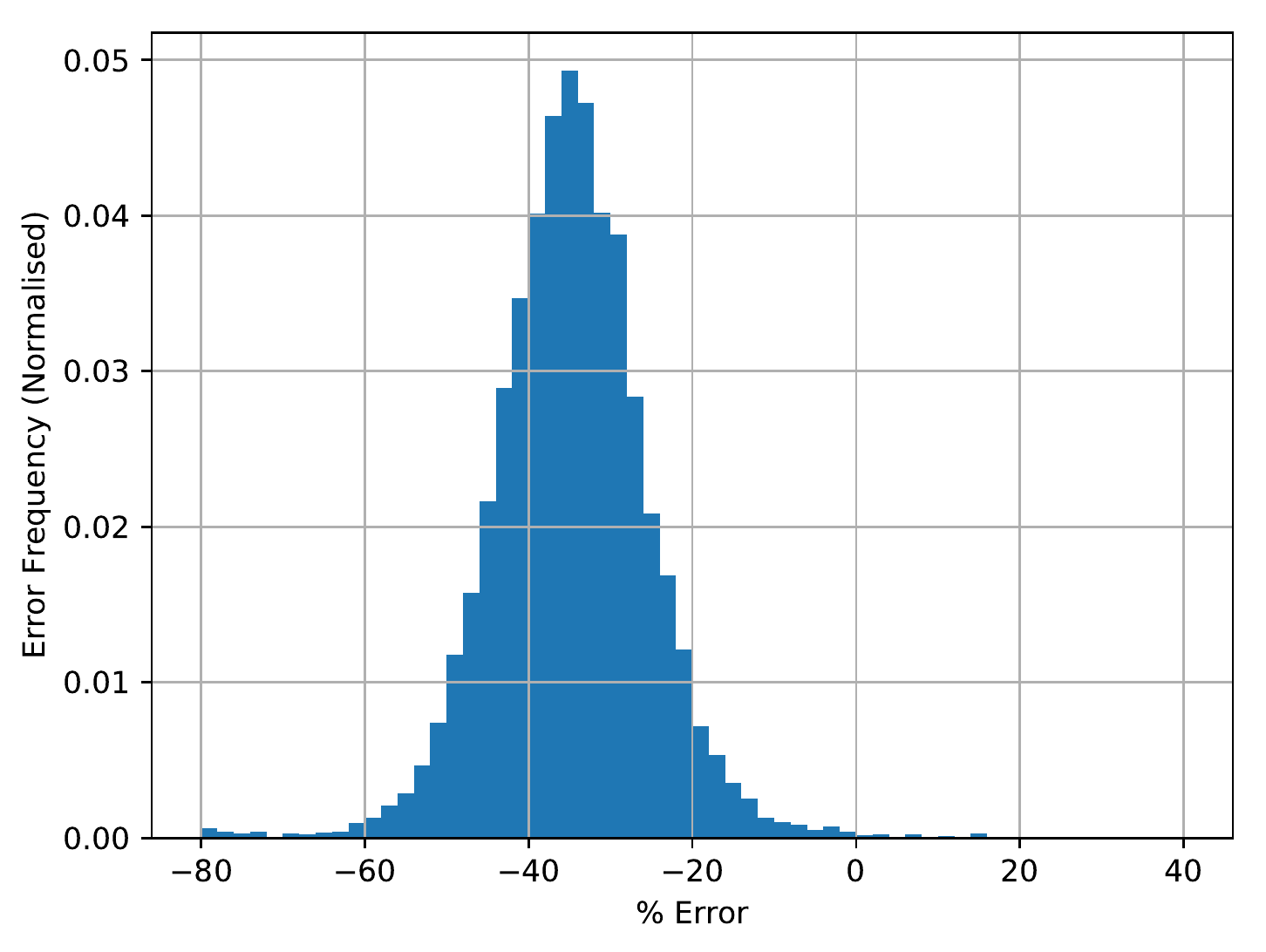} & \includegraphics[width=4cm]{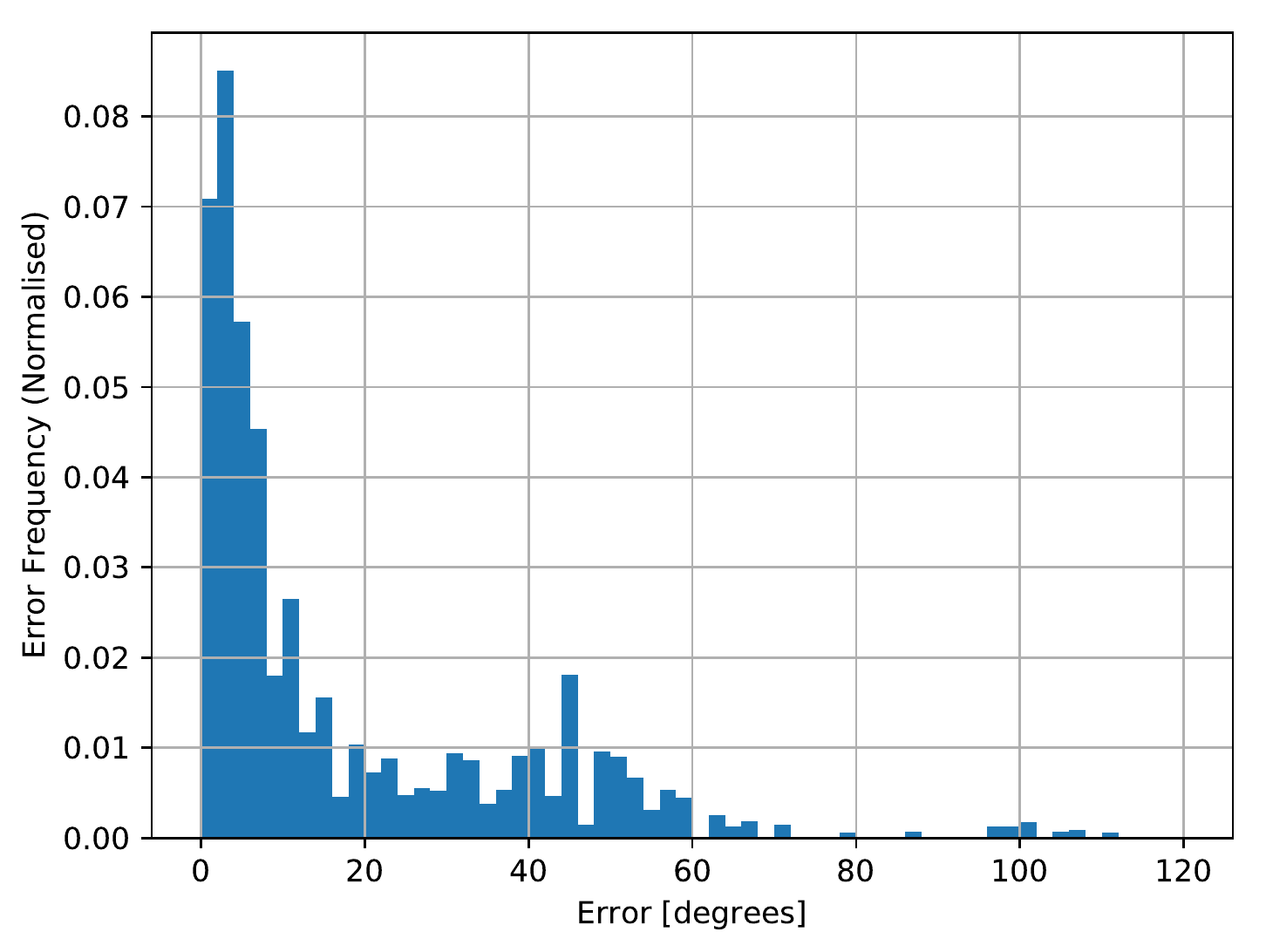} & \includegraphics[width=4cm]{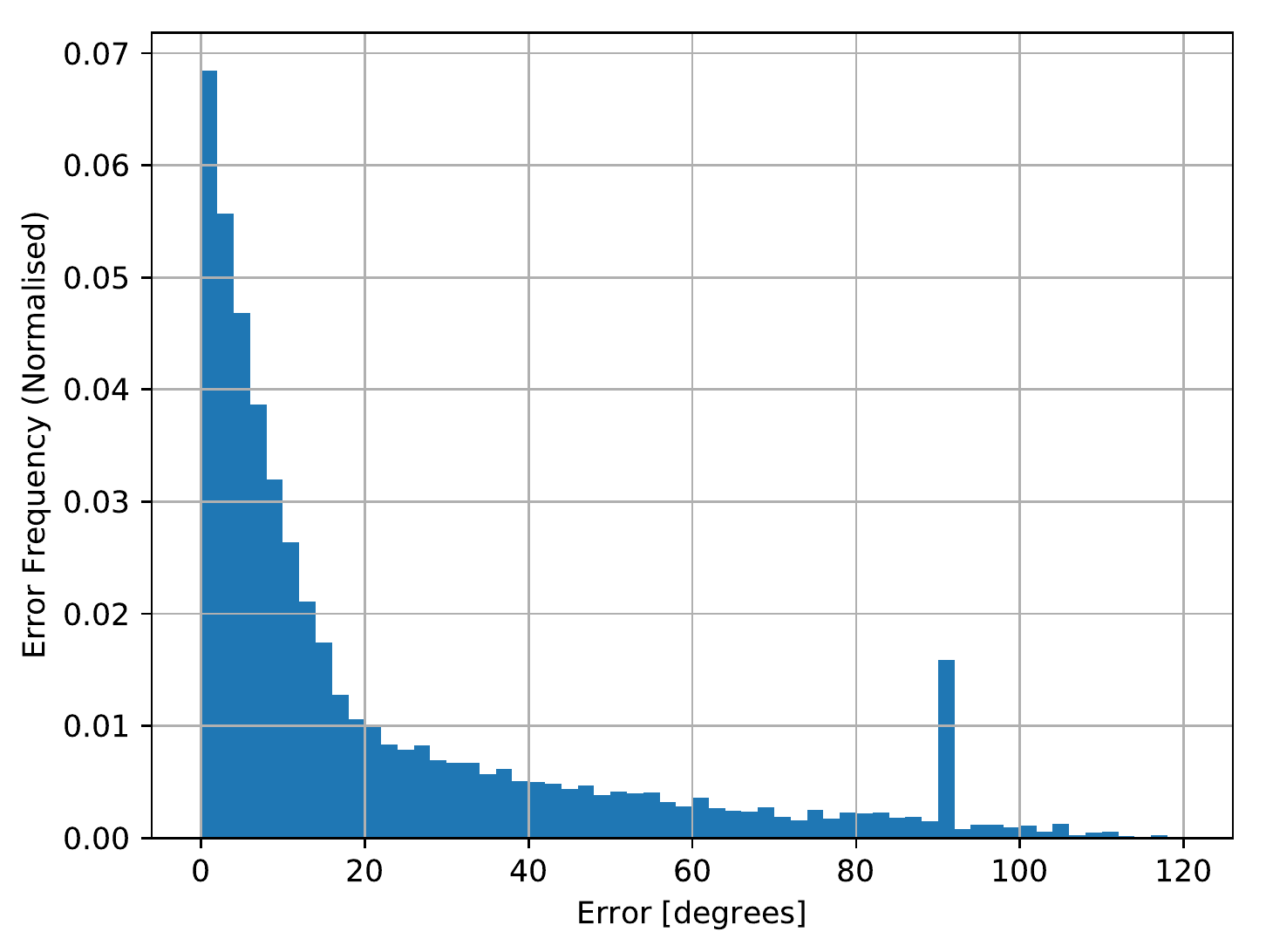} \\
            \bottomrule
        \end{tabular}
        \caption{Comparison to Lucas-Kanade via \% errors of flow vector magnitudes and errors of flow vector angles. Flow velocity = 58 pix/s}
        \label{fig:CompareLK}
    \end{figure*}
    \begin{figure*}[h]
        \centering
        \begin{tabular}{cM{45mm}M{45mm}M{45mm}}
           \toprule
           	Velocity & 5.8 pix/s & 58 pix/s & 289 pix/s\\
            \midrule
           	Magnitude & \includegraphics[width=4.5cm]{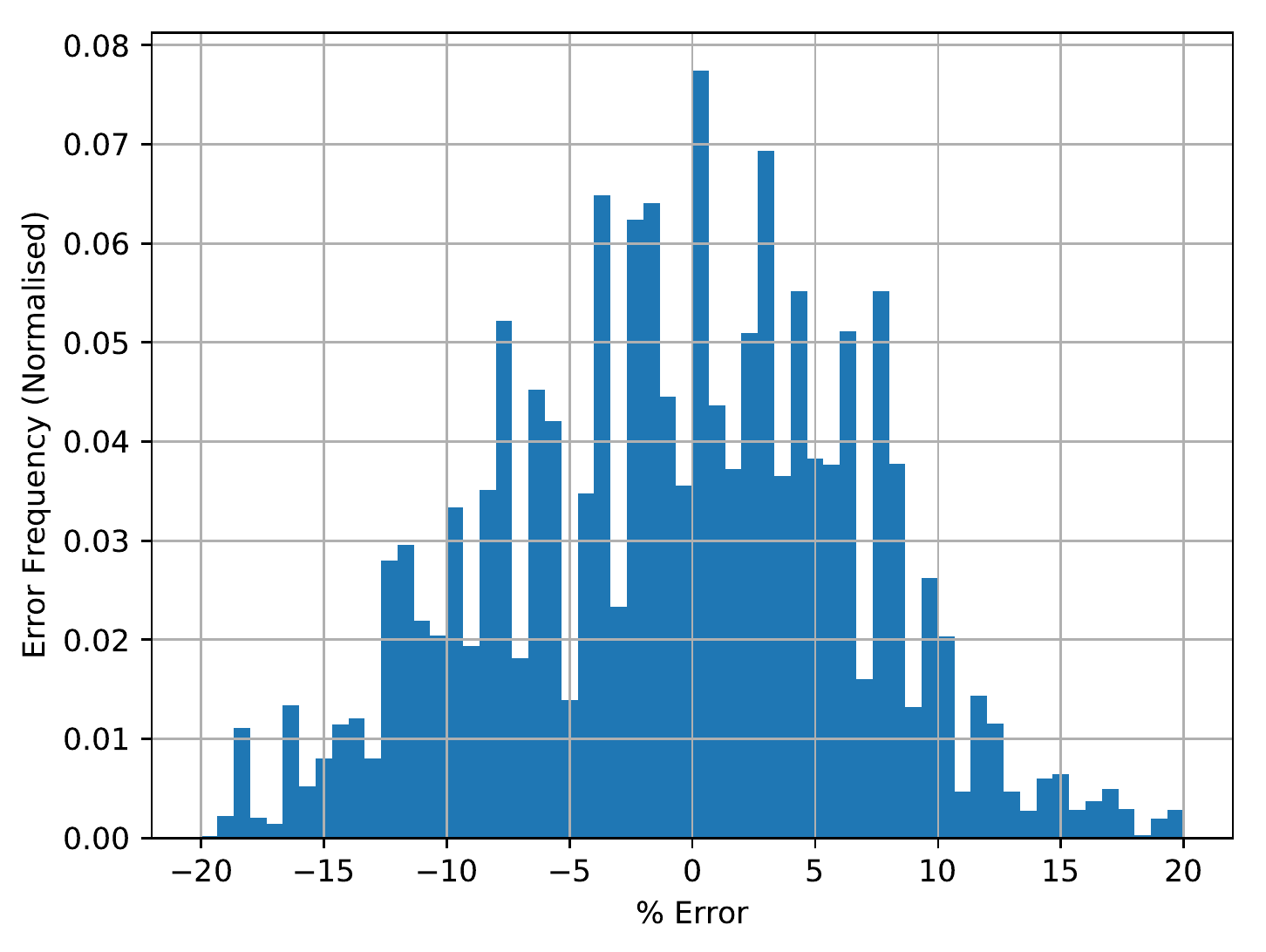} & \includegraphics[width=4.5cm]{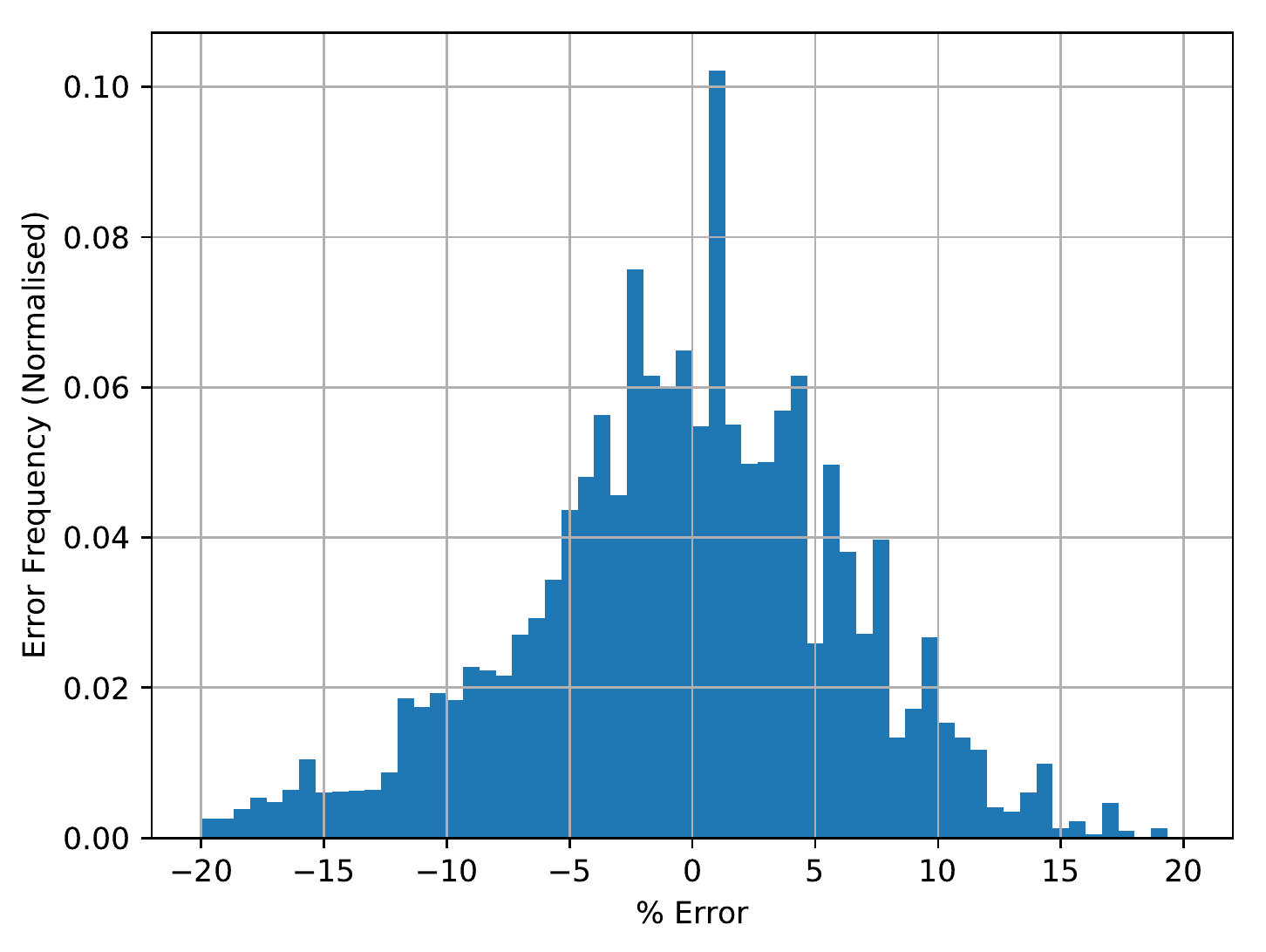} & \includegraphics[width=4.5cm]{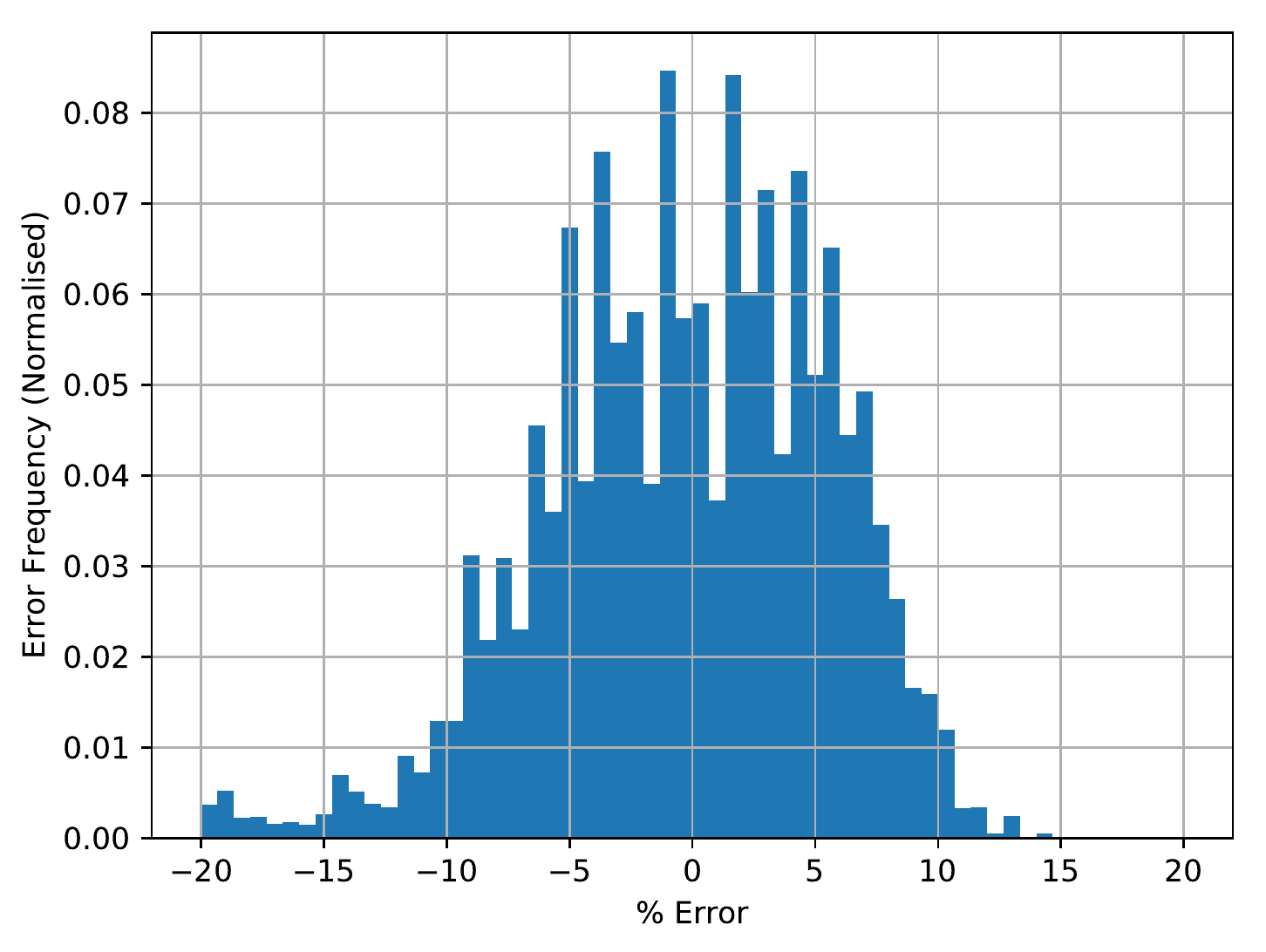} \\
           		Angle & \includegraphics[width=4.5cm]{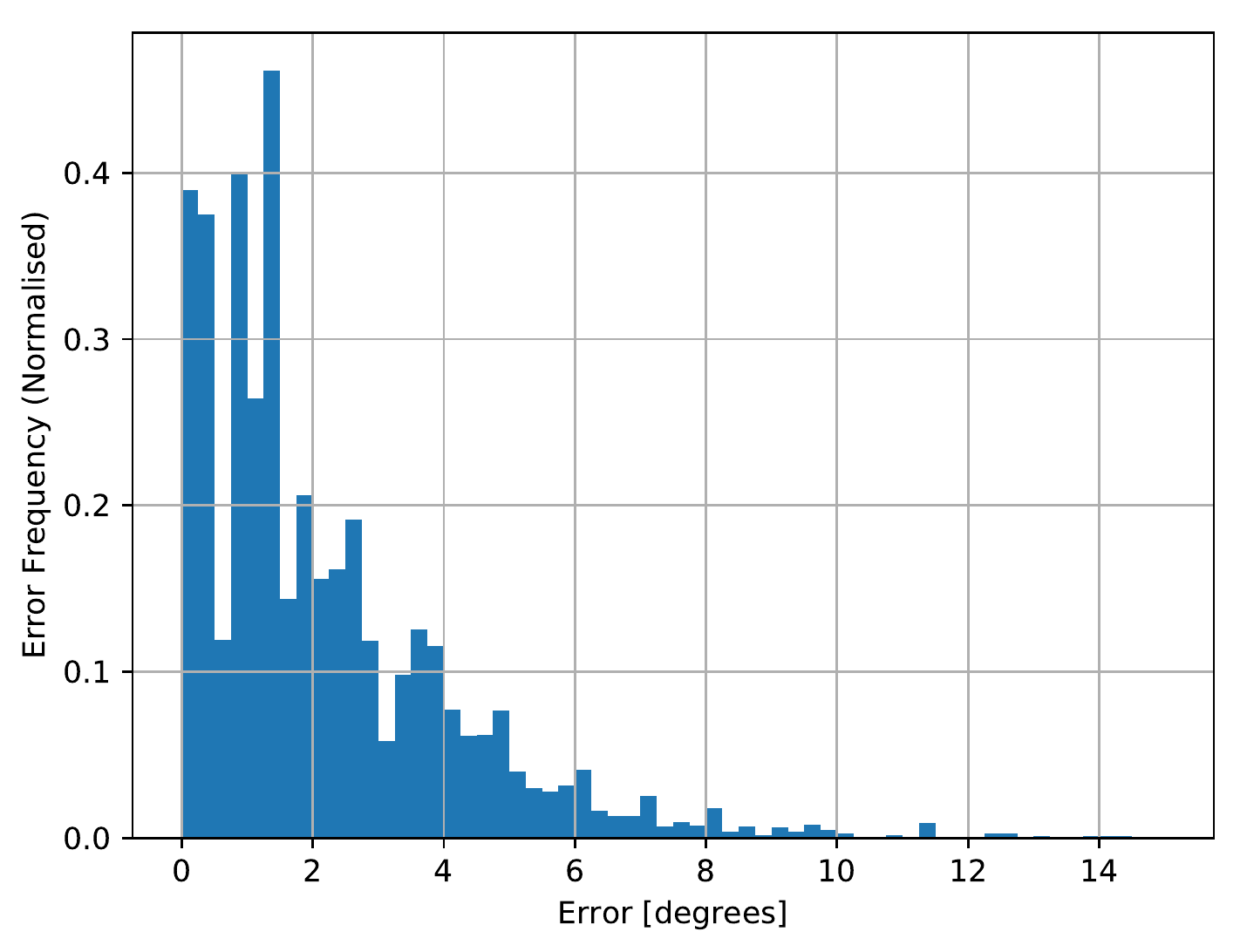} & \includegraphics[width=4.5cm]{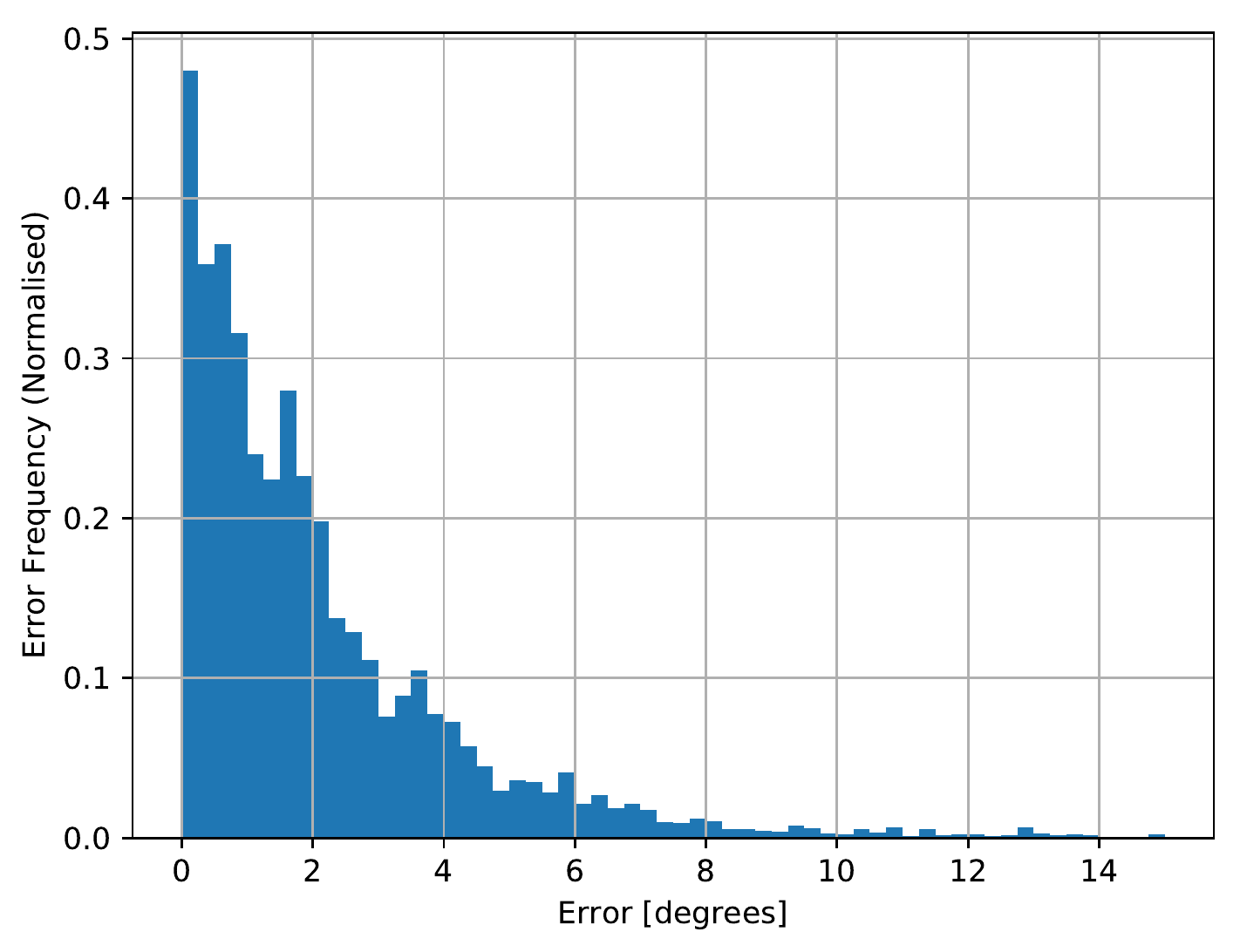} & \includegraphics[width=4.5cm]{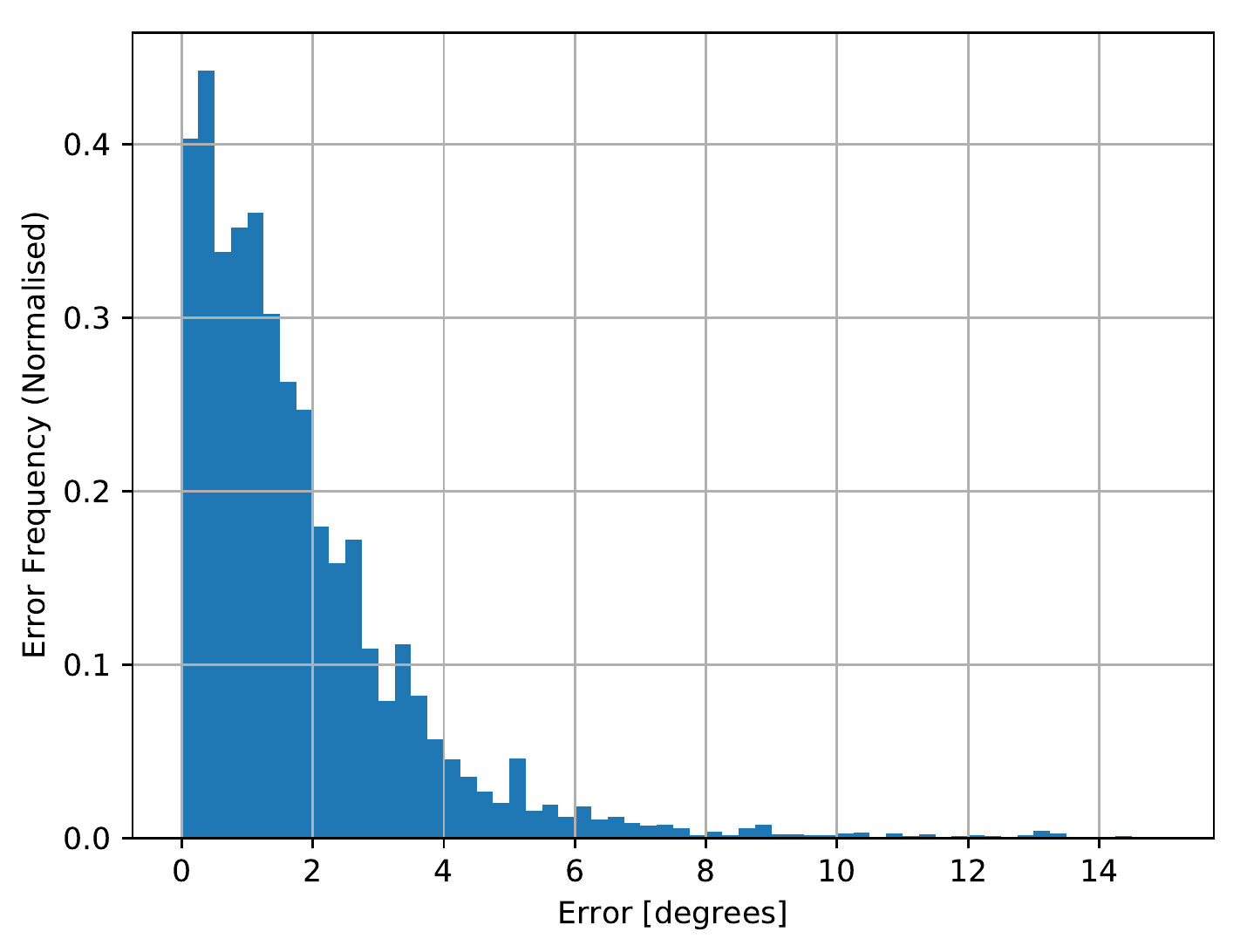}\\
            \bottomrule
        \end{tabular}
        \caption{\% error of flow vector magnitudes and flow vector angle error at a range of velocities computed using SOFAS}
        \label{tbl:CompareVel}
    \end{figure*}
The data used to validate the algorithm was generated with a DAVIS-240C DVS from Inilabs. The algorithm was written in Java using the jAER toolbox \cite{Delbruck2017jAER} and compared to an event based Lucas-Kanade implemented by Bodo Rueckauer \cite{Rueckauer2014Eval} and contained in the jAER toolbox. Since the algorithm performs several tasks, the tests are broken up into several sections. First the accuracy of the flow vectors estimated by algorithm is compared to that of Lucas-Kanade on a ground-truth dataset we generated. We chose to benchmark our algorithm against Lucas-Kanade since it returned the lowest errors of flow vectors of any of the optic flow implementations in the jAER toolbox. Having established the competitiveness of our algorithm in comparison to traditional optical flow estimation algorithms, we then examine the performance of the algorithm at different velocities and with a rapidly accelerating dataset when compared to ground truth. Having tested the accuracy, we then show the rate at which the estimate converges and finally we show qualitative results in more complex scenes for which we were unable to generate ground truth data, but which show the ability of the algorithm to successfully segment the scene into structures with distinct flow velocities.\par
Ground truth optical flow data was generated by using a UR5/CB3 robot arm \cite{UR52017Tech}. Magnets were mounted on the end effector of the robot, so that the robot was able to move simple structures with magnets attached across a background panel without generating events itself. Since the velocity of the robot end effector and the field of view of the DVS were known, the optical flow in the horizontal direction could be determined with ease (flow vector $\vec{v}=w_c\frac{\vec{v}_r}{w_{fov}}$) where $w_c$ is the width of the camera plane in pixels (240 on the DAVIS 240C), $\vec{v}_r$ is the end effector velocity and $w_{fov}$ is the width of the field of view.  
\subsection{Accuracy Comparison}
Here we contrast the percentage error to ground truth of the flow vector magnitudes and the error of the flow vector directions. The data used four different shapes with a flow velocity of 58 pix/s (a collection of rectangles (width=120 pixels), a hexagon (width=65 pixels), a large circle (width=70 pixels) and a small circle (width=7 pixels)). The results (Tab. \ref{fig:CompareLK}) show that our algorithm not only had a lower standard deviation of errors, but also was able to avoid the aperture problem. Results from Lukas-Kanade under-estimated the flow magnitude by about 35\%. This is because Lucas-Kanade, being a local estimation algorithm (in this case using a 7x7 pixel window), tends to estimate the flow vectors normal to the image gradients rather than in the direction of optical flow. Therefore, our method was also far superior in estimating the direction. The exception was for the small circle, where the methods were roughly even, since most of the object is able to fit into the kernel window in Lucas Kanade and the aperture problem is therefore avoided.
%Plots/TrackPlanes2/.pdf
\begin{figure}
    \centering
    \begin{subfigure}[b]{1\columnwidth}
        \includegraphics[width=\textwidth]{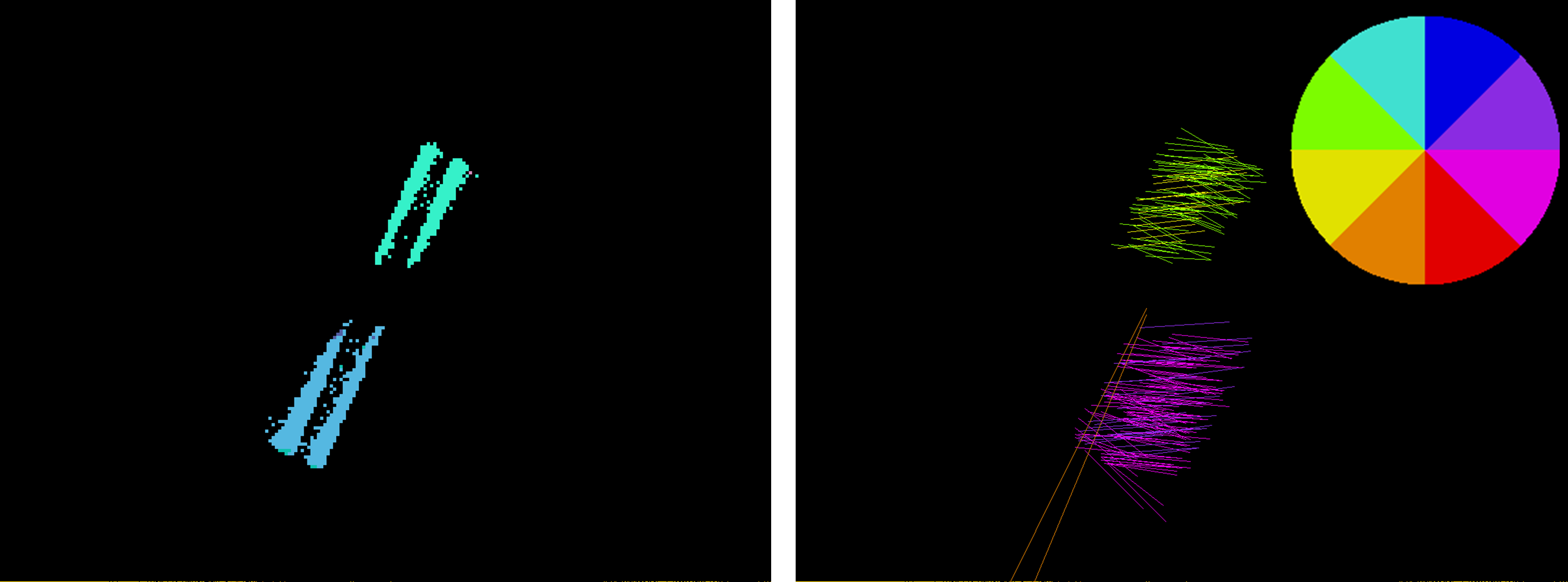}
        \caption{Short bar}
        \label{fig:shortbarRotating}
    \end{subfigure}
    \begin{subfigure}[b]{1\columnwidth}
        \includegraphics[width=\textwidth]{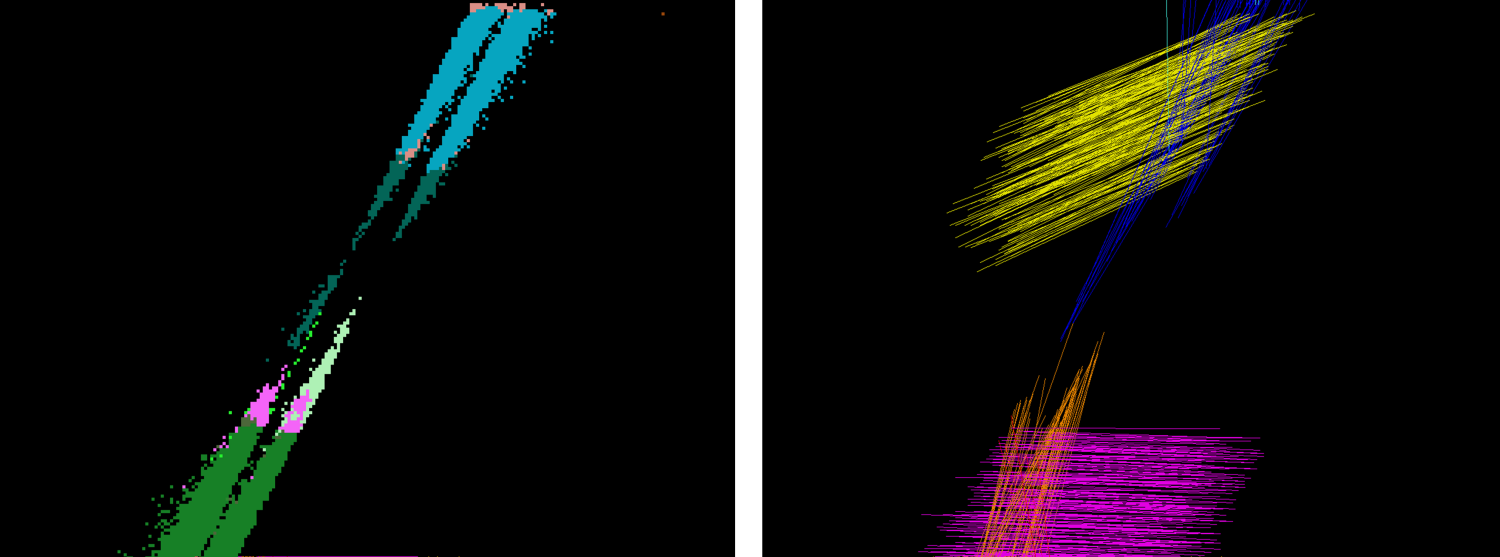}
        \caption{Long bar (filling entire frame)}
        \label{fig:longbarRotating}
    \end{subfigure}
    \caption{Two bars rotating counterclockwise at 1rps. On the left, the segmentation the algorithm performs, on the right the velocity vectors of the optic flow (see key in top right for disambiguating the direction)}
    \label{fig:barsRotating}
\end{figure}
\begin{figure*}
    \centering
    \begin{subfigure}[b]{1\textwidth}
        \includegraphics[width=\textwidth]{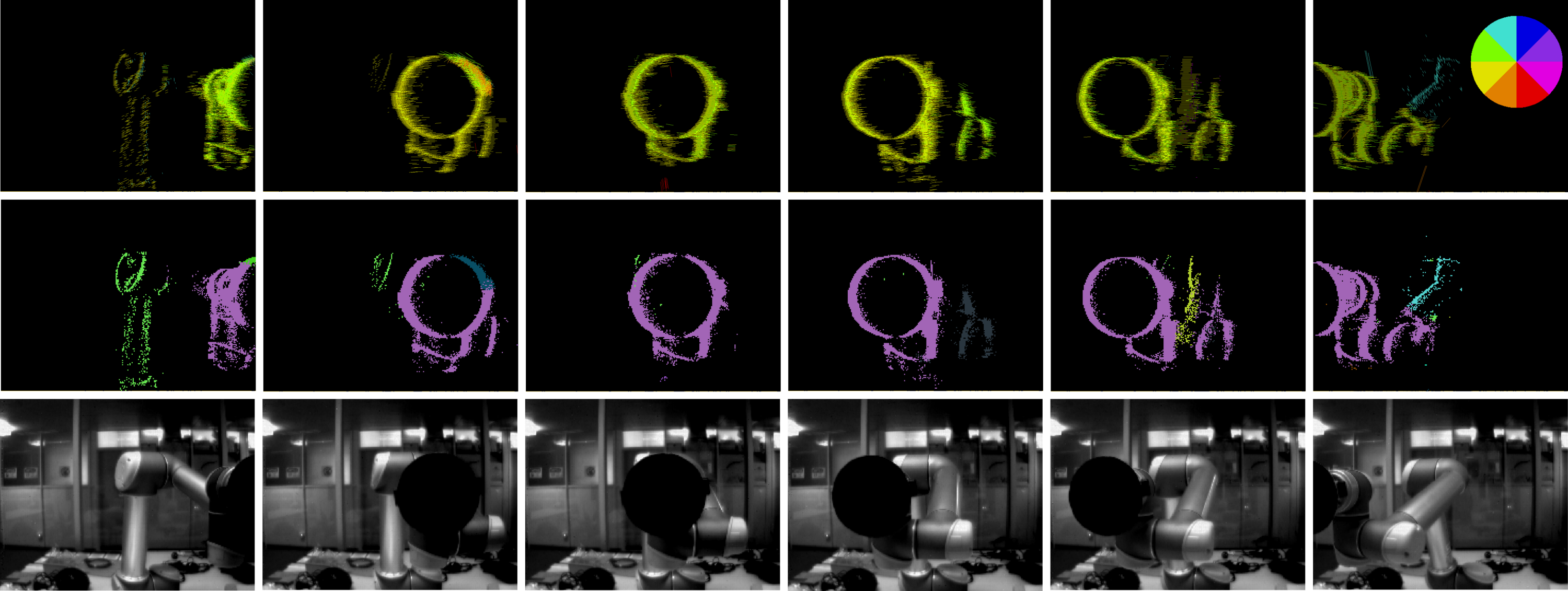}
        \caption{UR5 robot arm bearing a black cardboard circle moves the end effector from right to left in a horizontal, planar motion}
        \label{fig:Sequence01}
    \end{subfigure}
    \begin{subfigure}[b]{1\textwidth}
        \includegraphics[width=\textwidth]{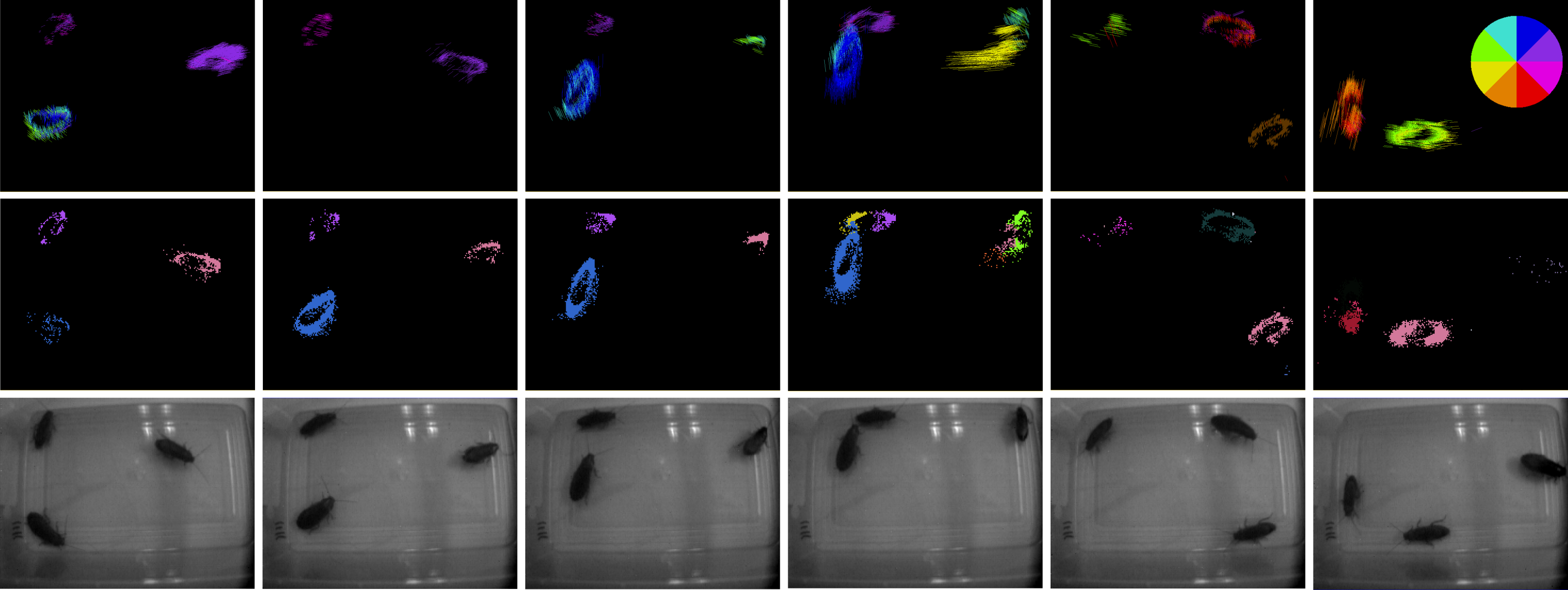}
        \caption{Cockroaches in a box scuttle around aimlessly}
        \label{fig:Sequence02}
    \end{subfigure}
    \begin{subfigure}[b]{1\textwidth}
        \includegraphics[width=\textwidth]{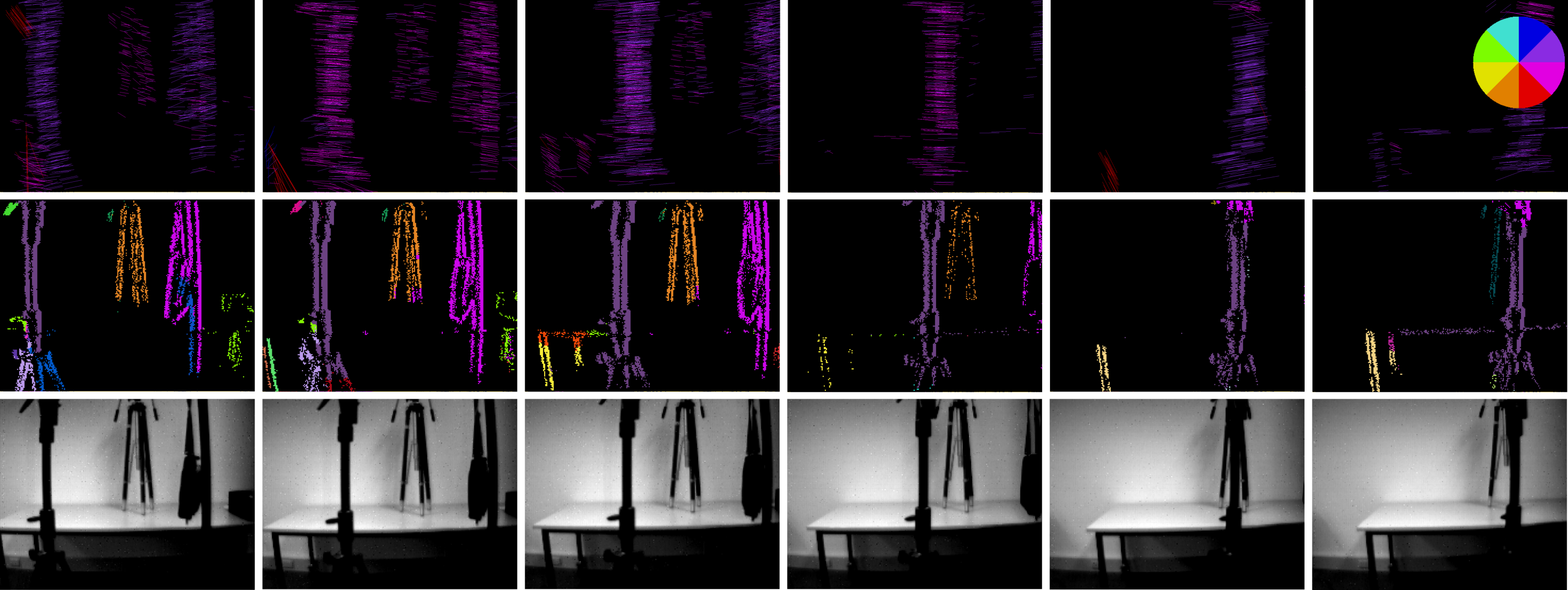}
        \caption{The camera pans past objects at different distances from the DVS. The tripod in the background (orange segmentation) is occluded and immediately recognised as a separate structure upon reappearance (green segmentation)}
        \label{fig:Sequence03}
    \end{subfigure}
    \caption{Snapshots of the algorithm. The top row shows the calculated flow vectors (see color wheel for direction), the middle row shows events coloured by segmentation, the bottom row shows grayscale stills of the scene}
    \label{fig:Sequences}
\end{figure*}
\subsection{Performance at Different Velocities}
Here we test our algorithm over a range of velocities using the large hexagon dataset. It shows that the algorithm performs slightly better at higher velocities. This is likely to do with the fact that the DVS used tends to generate more events over the same path at a higher velocity.
%\FloatBarrier
\subsection{Performance with Acceleration}
\begin{figure}
	\centering
    \includegraphics[width=\columnwidth]{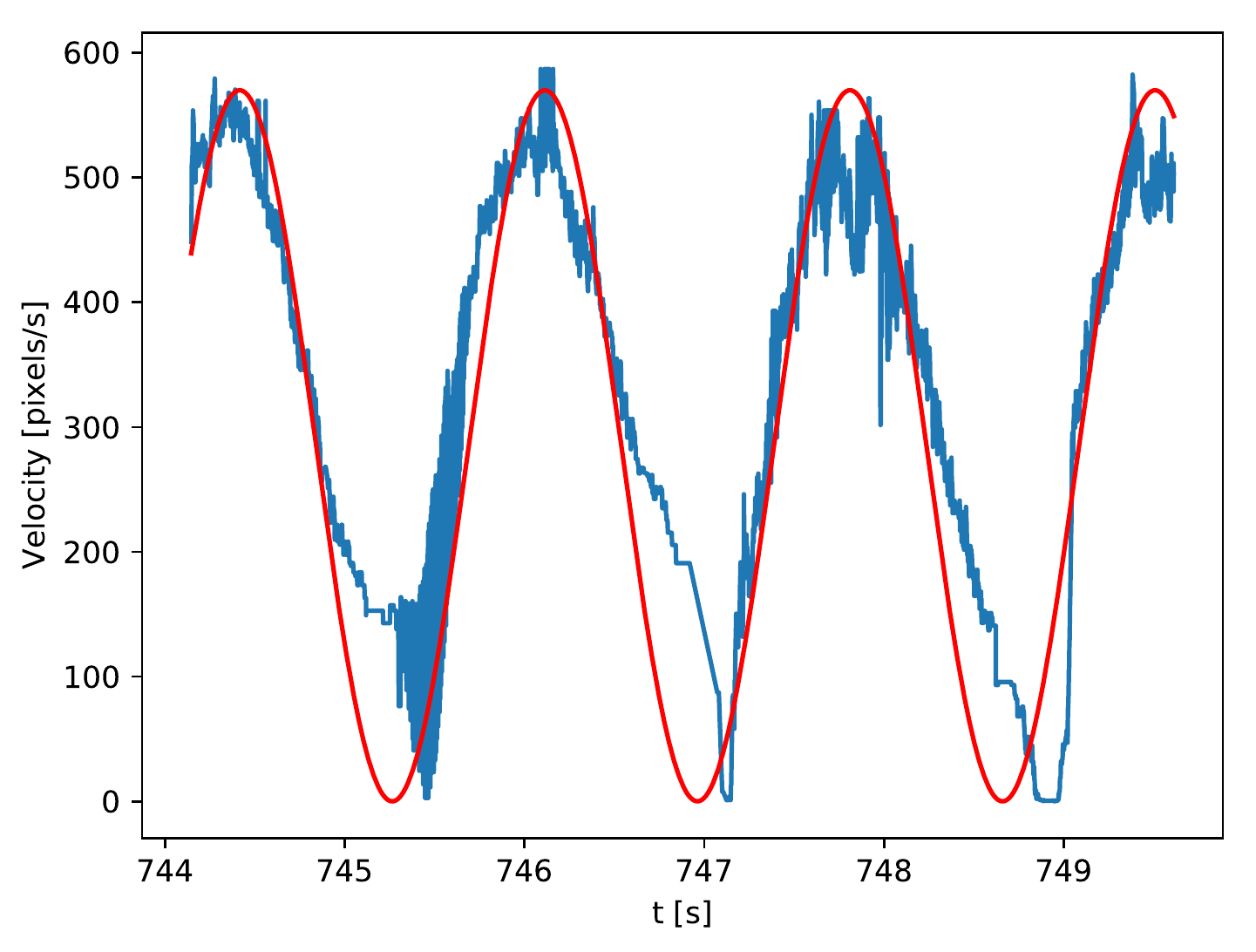}
     \caption{Estimated optic flow magnitude (blue) vs calculated ground truth (red) for the events generated by a pendulum (Length $L=0.72$m, starting angle $\theta_{max}=23^{\circ}$)}
    \label{fig:pendulum_vel4}
\end{figure}

The ability of the algorithm to deal with acceleration (which is not explicitly modelled in the projections, which assume locally constant velocity) was tested by comparing the computed flow magnitude of a pendulum to the calculated velocity. Since the maximum velocity of a pendulum is given by $v_{max}=\sqrt{2gL(1-cos(\theta_{max}))}$ and the period by $T=2\pi\sqrt{\frac{L}{g}}$  where $g=9.82ms^{-2}$, the pendulum length $L=0.72m$ and the maximum starting angle of the pendulum $\theta_{max}=23^{\circ}$, we were able to fully characterise the velocity and frequency of the pendulum, assuming a lossless sinusoidal motion model. Since the size of the field of view of the camera was also known, we were able to get ground truth for the optic flow magnitude. The phase shift was estimated from the footage. As can be seen in Fig. \ref{fig:pendulum_vel4}, the algorithm generates a relatively noisy estimate in cases where there are no constant velocities, but where the structures are under constant acceleration. Nevertheless, the algorithm is able to adapt to such circumstances to generate reasonable estimates.
\subsection{Rotation}
In this experiment, two bars were set to rotate in front of the DVS. Predictably, SOFAS performs poorly in such circumstances, since the projection does not model rotation (Fig. \ref{fig:barsRotating}). Especially when the rotating bar fills the entire screen, the algorithm suffers somewhat from the aperture problem as indicated by the grossly incorrect flow predictions near the centre (blue and orange vectors, Fig. \ref{fig:longbarRotating}). Nevertheless, the algorithm does attempt to segment the bar lengthwise and for the short bar at least generates reasonable optical flow. This is because over short enough timeframes it is able to assume the velocity as constant.
\subsection{Segmentation}
Here we aim to showcase the algorithms ability to segment the scene into structures with distinct velocities. To do this, we show the contours of the Track Planes generated on three datasets; one in which the UR5/CB3 robot arm moves the circle horizontally, one in which three cockroaches are moving randomly in a box and one in which the algorithm pans across a simple scene with four vertical bars at different distances to the DVS. As is clear (Fig. \ref{fig:Sequences}), SOFAS is able to clearly segment objects by their flow velocities, and event when the same object is segmented into multiple sections (see frame two of Fig. \ref{fig:Sequence01}), it is able to merge these together again. Because of the nature of the algorithm, there is some quantisation to the segmentation, though the extent of this is not easily quantifiable. In particular the result from Fig. \ref{fig:Sequence03} is important, since the effective segmentation of structures at different distances makes this algorithm a potential candidate for visual odometry in the future. Indeed, it should be noted how well the segmentation in Fig. \ref{fig:Sequence03} performs, separating the objects in the scene perfectly. When the tripod in the background is occluded, SOFAS is able to re-segment it as a structure immediately after reappearance. At the same time in this scene, the edges of the table are segmented into separate entities as they have distinct optic flow velocities.
\section{Conclusion and Future Work}
This paper presents an algorithm for detecting the optical flow of events generated by a DVS in a novel way that is better suited to the event-paradigm than traditional optical flow techniques. Because of the nature of the algorithm, segmentation by flow velocity comes ``free". In the course of this paper, SOFAS has been demonstrated to be robust and to be superior in accuracy when compared to a traditional optical flow algorithm.\par
We plan in future work to produce an optimised implementation of this algorithm. Since the computations involved are relatively basic and the overall structure of the algorithm (Fig. 5) is modular and inherently parallel, we feel that it should be possible to produce a real-time capable implementation. For the implementation presented here, the hexagon dataset from Fig. 8 required 13,602 ms to process (81,700 events) and the robot arm dataset from Fig. 11 took 46,461 ms to process (257,571 events). The algorithm was run on a PC with a 3.4 GHz CPU.\par
Clearly, fail cases for this algorithm are those in which most of the structures generating events on the image plane are accelerating or rotating rapidly. This problem could be solved, or at least improved by adding a dimension to the projection used, so that the acceleration is modelled also, a compelling direction for future work. Nevertheless, the adaptive nature of our algorithm enabled it to give reasonable estimates even under such non-ideal conditions. Other fail cases are likely to occur in situations where fast global lighting changes (such as a flickering light source) cause large numbers of events to be generated. This however, is an intrinsic weakness of DVSs and not a particular failing of our algorithm.
\subsection{Acknowledgements}
A short video showing the output of this algorithm can be found here: https://youtu.be/JVkQOW\_iUqs\par
This work was supported by the Australian Research Council Centre of Excellence for Robot Vision, project number CE140100016 (www.roboticvision.org).
\FloatBarrier
\balance
\bibliography{Bibliography} 
\end{document}